%% file: neurips_2026.tex
\documentclass{article}

\PassOptionsToPackage{numbers, compress}{natbib}

\usepackage[main,final]{neurips_2026}

\usepackage[utf8]{inputenc} 
\usepackage[T1]{fontenc}    
\usepackage{hyperref}       
\usepackage{url}            
\usepackage{microtype}      
\usepackage{graphicx}
\usepackage{subcaption}
\usepackage{booktabs}       
\usepackage{amsfonts}       
\usepackage{nicefrac}       
\usepackage{xcolor}         

\usepackage{amsthm}
\usepackage{amsmath}
\usepackage{amssymb}
\usepackage{mathtools}
\usepackage{amsthm}
\usepackage{algorithm}
\usepackage{algorithmic}
\usepackage{float}
\usepackage{bbm}
\usepackage{pifont}         
\usepackage{placeins}
\usepackage{wrapfig}
\usepackage[capitalize,noabbrev]{cleveref}

\theoremstyle{plain}
\newtheorem{theorem}{Theorem}[section]
\newtheorem{proposition}[theorem]{Proposition}

\theoremstyle{definition}

\theoremstyle{remark}

\newcommand{\Id}{\mathbbm{1}_\mathcal{C}}

\newcommand{\cmark}{\ding{51}}%
\newcommand{\xmark}{\ding{55}}%

\newcommand{\dd}{\mathrm{d}}

\usepackage[textsize=tiny]{todonotes}

\title{Predict-Project-Renoise: Sampling Diffusion Models under Hard Constraints}

\author{%
  Omer Rochman-Sharabi \\
  University of Liège\\  
  \texttt{o.rochman@uliege.be}
  \And
  Gilles Louppe \\
  University of Liège\\
}

\begin{document}

\maketitle

\begin{abstract}

Diffusion models cannot enforce hard constraints, yet applications in the physical sciences demand exact satisfaction of conservation laws, boundary conditions, and observational consistency. In this work, we identify a corrector kernel whose unique stationary distribution is the constrained marginal at each noise level, and approximate it by iteratively projecting through the denoiser and renoising via the forward kernel. The resulting Predict-Project-Renoise (PPR) algorithm enables sampling from pretrained diffusion models under hard constraints. Its three components are each necessary: projecting through the denoiser keeps samples close to the data manifold, while renoising and iterating drive samples toward the constrained marginal. On 2D distributions, the Kuramoto-Sivashinsky equation, and global weather forecasting with a $10^8$-dimensional atmospheric model, PPR simultaneously achieves low constraint violations and high distributional fidelity, a combination that existing methods fail to deliver.

\end{abstract}

 \section{Introduction}

Diffusion models \citep{sohldickstein2015deepunsupervisedlearningusing, ddpm, song2021scorebasedgenerativemodelingstochastic} have emerged as the state-of-the-art framework for deep generative modeling, showing strong performance in approximating high-dimensional data distributions $p$ across modalities. However, in many scientific and engineering domains, such as medical imaging, fluid dynamics, or climate modeling, sampling $p$ is insufficient. Samples $x$ must also satisfy hard constraints $x \in \mathcal{C}$, such as exact consistency with observations or governing equations, zero divergence in incompressible fluids, conservation laws, boundary and initial conditions, or geometric constraints.

Although training-based approaches can incorporate constraints \citep{zhouGeneratingPhysicalDynamics2025, khalafi2024constraineddiffusionmodelsdual, naderiparizi2025constrainedgenerativemodelingmanually}, they often lack flexibility, requiring retraining for different constraints, and enforce the constraints only softly. For these reasons, recent efforts focus on \emph{training-free constrained sampling}, where a pre-trained unconditional model is guided to satisfy constraints during inference. Current methods typically rely on treating constraints as observations \citep{dhariwal2021diffusionmodelsbeatgans} or implement projection-based corrections \citep{chung2024improving, rochman2025}.

When treating the constraint as an observation, Bayesian decomposition can be used to express the posterior score as a prior score plus a likelihood score. While effective for noisy inverse problems  \citep{daras2024surveydiffusionmodelsinverse}, especially in the context of images, these methods are ill-suited for hard constraints: the implicit likelihood approaches a Dirac delta, causing gradients to become unstable and informative only in the immediate vicinity of the feasible set $\mathcal{C}$. Projection methods can enforce feasibility explicitly, but have theoretical and practical pitfalls as they do not correctly sample the true constrained distribution. This manifests as three problems: (i) the constraints are violated, (ii) the constrained distribution is incorrect, and (iii) the sample lies outside of the support $\mathcal{S}$ of the data distribution $p$.

For nonlinear constraints, the projection typically has no closed-form solution, so it must be approximated with an iterative solver. Moreover, it is often unclear what exactly should be projected. Directly projecting the noisy state onto the constraint set can be ill-posed or numerically unstable, since many constraints are only meaningful for clean data. Projecting an estimate of the clean sample is not a fix either. Sample consistency with $p$ is often broken, for example, by introducing discontinuities into otherwise continuous samples. Finally, repeated projection steps can systematically bias the sampler. In the extreme, this behaves like an attractor toward ``nearest'' feasible points, causing mode collapse rather than sampling from the constrained distribution.

\paragraph{Contributions.}
In this work, we address these limitations by arguing that correct constrained sampling requires approximating the constrained marginal distributions $p_t^\mathcal{C}$ induced by a forward diffusion SDE starting from the constrained set $\mathcal{C}$. We define the constrained forward process and derive the associated constrained marginals and backward process. Based on the backward process, we identify a corrector kernel that has the constrained marginals as its unique stationary distributions, design the Predict-Project-Renoise (PPR) algorithm that approximates that kernel, and show that each of its components is necessary. PPR achieves higher feasibility rates, lower constraint violations, and better mass distribution compared to other methods.

\section{Score-based generative modeling}

\paragraph{Forward process.} Generative modeling seeks to approximate a data distribution $p(x)$ of $x \in \mathbb{R}^D$. In score-based diffusion \citep{song2021scorebasedgenerativemodelingstochastic}, this is done by learning to reverse a forward corruption process. Following \citet{lai2025principlesdiffusionmodels}, we define the \textit{forward process} as the stochastic differential equation (SDE)
\begin{equation} \label{eq:forward-diffusion}
    \dd x_t = f(x_t, t) \dd t + g(t) \dd w_t ,
\end{equation}

where $f: \mathbb{R}^{D+1} \to \mathbb{R}^D$ is a drift term, $g: \mathbb{R} \to \mathbb{R}$ is a scalar diffusion coefficient, and $w_t$ is a standard Wiener process. Starting from $p_0 = p$, this SDE defines a series of marginal densities $\{p_t(x_t)\}_{t \in [0,T]}$ which interpolate between the data distribution $p_0$ and $p_T$. Choosing $f$ and $g$ defines the perturbation or \textit{forward kernels} $p(x_t | x_0)$, inducing the marginals
\begin{equation} \label{eq:marginals}
    p_t(x_t) = \int p(x_t \mid x_0) \, p(x_0) \, \mathrm{d}x_0.
\end{equation}
If $f$ and $g$ are chosen appropriately, then $p(x_t | x_0)$ and $p_T$ become Gaussian \citep{sdebook}. In particular, $p_T$ can become $\mathcal{N}(0, \sigma_TI)$.

\paragraph{Backward process.} \citet{anderson82} shows that given a forward process (Eq.~\ref{eq:forward-diffusion}), there exists a \textit{reverse} or \textit{backward process} from $T$ to $0$ transporting $p_T$ back to $p_0$ given by
\begin{equation}  \label{eq:reverse-process}
    \dd x_t = \left[ f(x_t, t) - g^2(t) \nabla_{x_t} \log p_t(x_t) \right] \dd t + g(t) \dd \tilde{w}_t,
\end{equation}
where $x_T \sim p_T(x_T)$ and $\tilde{w}_t$ is the reverse-time Wiener process. In this reverse process, the marginal distributions of $x_t$ are identical to those of the forward process. This allows for sampling from $p$ by generating a sample from $p_T$ and then simulating the reverse process.

\paragraph{Training and sampling.}
Since the score $s(x_t) = \nabla_{x_t}\log p_t(x_t)$  is unknown, it has to be learned from data.
We follow \citet{elucidating} and parameterize the score using Tweedie's formula  $s_\theta(x_t) = (d_\theta(x_t) - x_t) / \sigma_t^2$ \citep{Tweedie1947Functions, Efron2011Tweedies}. The resulting denoising objective minimizes
\begin{equation}
    \mathbb{E}_{x_0 \sim p(x_0)} \mathbb{E}_{x_t \sim p(x_t | x_0)} \| d_\theta(x_t) - x_0 \|^2.
\end{equation}

 The denoiser $d_\theta(x_t) \approx \mathbb{E}[x_0 | x_t]$ predicts the posterior mean of the clean sample given the noisy input. Once it is available, a way to sample from the reverse process is to use the predictor-corrector (PC) sampler (\citet{song2021scorebasedgenerativemodelingstochastic}, Algorithm~\ref{alg:reverse-sampling}). This algorithm alternates between a \textsc{Predict} step (e.g. DDPM \citep{ddpm}, DDIM \citep{ddim}), which moves the samples from $p_{t+1}$ to $p_t$, and a number of \textsc{Correct} steps (e.g. Langevin steps based on the score). The \textsc{Correct} steps reduce the discretization error by pulling samples toward higher-density regions of $p_t$ and re-injecting noise to increase diversity. \textsc{Predict} can be any valid transition. In our experiments it is DDIM, which takes the denoised estimate $\hat{x}_0 = d_\theta(x_{t+1})$ as input.

\section{Method}

The reverse process (Eq.~\ref{eq:reverse-process}) allows us to approximate the data distribution $p$, but in many domains we are interested in imposing constraints on the samples. Let $\mathcal{S} \subseteq \mathbb{R}^D$ be the support of $p(x)$ and $\mathcal{C} \subseteq \mathbb{R}^D$ be a (constrained or feasible) set corresponding to samples with certain characteristics, such as conservation laws, adherence to given dynamics, or a desired decay in the energy spectrum. $\mathcal{C}$ can be represented using a nonnegative constraint function $c(x) \ge 0$ as $\mathcal{C} \coloneqq \{x \in \mathbb{R}^D : c(x) = 0\}$. The intersection $\mathcal{S} \cap \mathcal{C}$ strictly restricts the prior to the feasible set, inducing the density $p^\mathcal{C} \propto p(x_0) \Id(x_0)$, where $\Id(x)$ is the indicator function of $\mathcal{C}$. We say a sample $x \in \mathcal{S} \cap \mathcal{C}$ is \emph{feasible}.

\subsection{Constrained diffusion}
\label{sec:constrained-diffusion}

\paragraph{Constrained forward process.} We define the \textit{constrained forward process} similarly to Eq.~\ref{eq:forward-diffusion}, but with $x_0 \sim p^\mathcal{C}(x_0)$. This constrained process induces a sequence of marginals $\{p_t^\mathcal{C}\}_{t \in [0,T]}$, given by
\begin{equation} \label{eq:constrained-marginals}
    p_t^\mathcal{C}(x_t)
   \propto \int p(x_t \mid x_0) \, p(x_0) \, \Id(x_0) \, \mathrm{d}x_0.
\end{equation}
The unconstrained and constrained forward processes share the same dynamics and transition kernels but their different initial densities lead to distinct marginals and reverse processes.

\paragraph{Constrained backward process.} Mirroring the unconstrained backward process in Eq.~\ref{eq:reverse-process}, the \textit{constrained reverse process} is
\begin{equation} \label{eq:constrained-reverse}
    \dd x_t = \left[ f(x_t, t) - g^2(t) \nabla_{x_t} \log p_t^\mathcal{C}(x_t) \right] \dd t + g(t) \dd \tilde{w}_t.
\end{equation}

\begin{figure}
    \centering
    \includegraphics[width=\linewidth]{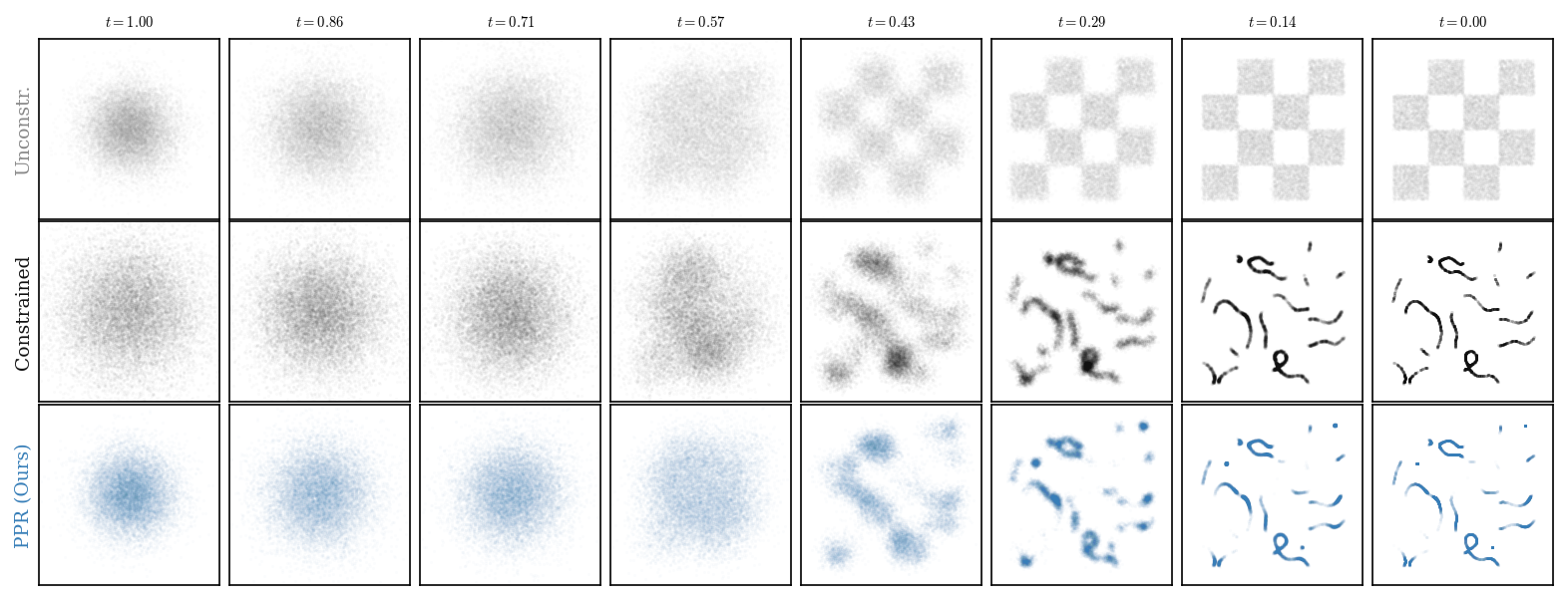}
    \caption{Visualization of the constrained process on \textsc{Data2D}. The top row shows the unconstrained marginals $p_t$. The middle row shows the target constrained marginals $p_t^\mathcal{C}$, obtained by applying the diffusion transition kernel to $p_0^\mathcal{C}$, which was sampled by rejection sampling. The bottom row shows the marginals induced by PPR. The columns progress in diffusion time, with the last column corresponding to $t=0$ (clean data).}
    \label{fig:toy-clouds}
\end{figure}

This process involves the constrained score function $ \nabla_{x_t} \log p_t^\mathcal{C}(x_t)$. It can be computed by taking the log and then the gradient of the constrained marginals (Eq.~\ref{eq:constrained-marginals})
\begin{equation}
    \nabla_{x_t} \log \int_{\mathcal{C}} p(x_t | x_0) p(x_0) \dd x_0,
\end{equation}
dropping the normalization constant because it vanishes under the derivative and absorbing the indicator function into the integral. Then, from Bayes' theorem, $p(x_t|x_0)p(x_0) = p(x_0|x_t)p(x_t)$, we get
\begin{equation} \label{eq:constrained-score}
    \nabla_{x_t} \log p_t^\mathcal{C}(x_t) =  \nabla_{x_t} \log p_t(x_t) + \nabla_{x_t} \log \int_{\mathcal{C}} p(x_0 | x_t) \dd x_0.
\end{equation}
The constrained score is thus the unconstrained score plus the gradient of the log-feasibility $ \nabla_{x_t} \log \int_{\mathcal{C}} p(x_0 | x_t) \dd x_0 = \nabla_{x_t} \log  \mathbb{P}(x_0 \in \mathcal{C} |x_t)$, a term pointing in the direction (in $x_t$-space) of increasing probability of $x_0$ being in $\mathcal{C}$. Figure~\ref{fig:toy-clouds} gives visual intuition on these processes. No additional training is required. Eqs.~\ref{eq:constrained-marginals} and~\ref{eq:constrained-reverse} define the target processes analytically, and PPR imposes the constraint at inference time on a pretrained denoiser, which can be reused for different constraints.

\subsection{Predict-Project-Renoise (PPR) sampling}

At each reverse step, the unconstrained predictor (Algorithm~\ref{alg:reverse-sampling}) provides $x_t \sim p_t$, whose denoised estimates are unlikely to satisfy the constraints exactly. Based on the structure of the constrained backward process (Section~\ref{sec:constrained-diffusion}), we identify the following corrector:

\begin{proposition} \label{prop:station}
Let $K_t(x_t' | x_t) = \int_\mathcal{C} p(x_t'|x_0) \, p^\mathcal{C}(x_0|x_t) \, dx_0$ with $p^\mathcal{C}(x_0|x_t) = \frac{p(x_0|x_t) \, \Id(x_0)}{\mathbb{P}(x_0 \in \mathcal{C}|x_t)}.$ Under this kernel, $p_t^\mathcal{C}$ is the unique stationary distribution and $(K_t)^M q \to p_t^\mathcal{C}$ as $M \to \infty$ for any initial distribution $q$. See Appendix~\ref{app:thm} for the proof.
\end{proposition}

That is, iterating the kernel $K_t$ at a fixed noise level acts as a corrector that drives any sample, including $x_t \sim p_t$ from the predictor, toward the constrained marginal $p_t^\mathcal{C}$, just as Langevin correction drives samples toward $p_t$. Its structure is a two-step conditional resampling: sample a feasible $x_0 \in \mathcal{C}$ from the constrained posterior $p^\mathcal{C}(x_0|x_t)$, then renoise via the forward kernel to obtain $x_t'$. Renoising is essential as it restores the correct noise level and injects the stochasticity needed for mixing.

Of course, sampling from the constrained posterior $p^\mathcal{C}(x_0|x_t)$ is itself intractable. Instead, PPR approximates it with a point estimate $p^\mathcal{C}(x_0|x_t) \approx \delta(x_0 - d_\theta(\textsc{Project}(x_t)))$, 
where $\textsc{Project}(x_t) \coloneqq \arg\min_x c(d_\theta(x))$ is solved via an off-the-shelf optimizer initialized at $x_t$. 
Plugging this approximation into $K_t$ yields
\begin{equation}
    K_t^{\mathrm{PPR}}(x_t'|x_t) \coloneqq p\left(x_t' \mid d_\theta (\textsc{Project}(x_t)) \right),
\end{equation}
where the integral over $\mathcal{C}$ in $K_t$ collapses to a single forward kernel evaluation at the projected point estimate. Again, the renoising step is critical: without it, $K_t^{\mathrm{PPR}}$ reduces to a deterministic map that collapses to fixed points of the projection; with it, the kernel retains the stochasticity needed for mixing. Following Proposition~\ref{prop:station}, we iterate $K_t^{\mathrm{PPR}}$ $M$ times to drive samples toward $p_t^\mathcal{C}$. In summary, the proposed \emph{Predict-Project-Renoise} sampler (Algorithm~\ref{alg:general_projection}) alternates at each reverse step between a predictor update, a projection through the denoiser, and a renoising step, with the last two repeated $M$ times. 

\paragraph{Approximation error.} Replacing the constrained posterior with a point estimate introduces an error that we characterize in Appendix~\ref{app:thm}. The error decomposes into two terms: a mixing term that decays exponentially with $M$, justifying the repetition of project-renoise steps, and an irreducible bias from the point estimate approximation, which is small when the posterior concentrates around a single feasible $x_0$. Proposition~\ref{prop:station} concerns the ideal kernel. Practical PPR replaces the conditional draw with a projected point estimate, and a finite optimizer does not solve every nonconvex $c$ exactly. Feasibility is therefore evaluated at a stated tolerance.

\paragraph{Projecting through the denoiser.}
There are three natural ways to enforce feasibility during reverse sampling. The first is to minimize $c(x_t)$ with respect to $x_t$ directly. However, this is problematic because constraints are defined on clean data, while $x_t \notin \mathcal{S}$ almost surely, making $c(x_t)$ ill-conditioned or undefined. The second is to compute $x_0 = d_\theta(x_t)$ and minimize $c(x_0)$ with respect to $x_0$. This evaluates the constraint where it is meaningful, but the optimized $x_0$ is free to leave $\mathcal{S}$, producing artifacts such as discontinuities (Section~\ref{sec:experiment}). PPR takes a third route: minimize $c(d_\theta(x_t))$ with respect to $x_t$. Because the denoiser is trained to output elements of $\mathcal{S}$, the projected estimate $d_\theta(x_t^*)$ is encouraged to remain on the data manifold. Backpropagation through $d_\theta$ further propagates local constraint information through learned correlations, helping maintain $x_0 \in \mathcal{C} \cap \mathcal{S}$ globally. The cost is higher due to differentiation through $d_\theta$, but can be negligible when the constraint itself dominates computation, as in latent-space models with large decoders. In practice, \textsc{Project} is solved with off-the-shelf optimizers, L-BFGS~\citep{lbfgs} for \textsc{Data2D} and \textsc{KS} and Adam~\citep{adam} for \textsc{Weather}, with a fixed maximum of $8$ iterations and no convergence-based stopping. We observed no NaNs, failed solves, or need for restarts. Renoising perturbs the state after a possibly imperfect solve, so later \textsc{Project} applications do not repeat the same optimization trajectory. The projection budget provides a direct tradeoff between compute and sample feasibility.

\begin{figure}[b]
  \centering
  \begin{minipage}[t]{0.49\linewidth}
    \vspace{-0.5cm}
    \begin{algorithm}[H]
      \caption{Predictor-corrector sampling.}
      \label{alg:reverse-sampling}
      \begin{algorithmic}[1]
        \STATE \textbf{Input:} denoiser $d_{\theta}$, predictor $\textsc{Predict}$, corrector $\textsc{Correct}$, and $N$.
        \STATE \textbf{Initialize:} sample $x_T \sim p_T(x_T)$
        \FOR{decreasing $t \in \{T-1, \dots, 0\}$}
          \STATE $\hat{x}_0 \leftarrow d_{\theta}(x_{t+1})$
          \STATE $x_t \leftarrow \textsc{Predict}\!\left(x_{t+1}, \hat{x}_0, t+1 \right)$
          \FOR{$i = 1, \dots, N$}
            \STATE $\hat{x}_0 \leftarrow d_{\theta}(x_t)$
            \STATE $x_t \leftarrow \textsc{Correct}(x_t, \hat{x}_0, t)$
          \ENDFOR
        \ENDFOR
        \STATE \textbf{Output:} $x_0$
      \end{algorithmic}
    \end{algorithm}
  \end{minipage}\hfill
  \begin{minipage}[t]{0.49\linewidth}
    \vspace{-0.5cm}
    \begin{algorithm}[H]
      \caption{Predict-Project-Renoise (PPR).}
      \label{alg:general_projection}
      \begin{algorithmic}[1]
        \STATE \textbf{Input:} denoiser $d_{\theta}$, predictor \textsc{Predict}, projection $\Pi_d$, steps $M$.
        \STATE \textbf{Initialize:} sample $x_T \sim p_T(x_T)$
        \FOR{decreasing $t \in \{T-1, \dots, 0\}$}
          \STATE $\hat{x}_0 \leftarrow d_{\theta}(x_{t+1})$
          \STATE $x_t \leftarrow \textsc{Predict}\!\left(x_{t+1}, \hat{x}_0, t+1 \right)$
          \FOR{$i = 1, \dots, M$}
            \STATE \textcolor{red}{$x_t^* \leftarrow \textsc{Project}(x_t)$}
            \STATE \textcolor{red}{$x_t \sim p(x_t \mid x_0 = d_\theta(x_t^*))$} \COMMENT{\textsc{Renoise}}
          \ENDFOR
        \ENDFOR
        \STATE \textbf{Output:} $x_0$
      \end{algorithmic}
    \end{algorithm}
  \end{minipage}
\end{figure}

\section{Related Work} \label{sec:related}

Previous work on constrained sampling in diffusion models is summarized in Table~\ref{tab:algo_mapping}. First, we focus on projection-based constrained samplers and then discuss a related body of work on posterior sampling under noisy observation processes. To help understand how previous methods relate to each other, we annotate each projection-based method according to the space in which the projection is done. If done with respect to $x_t$, we say the projection is in \textcolor{red}{$x_t$-space} and use \textcolor{red}{red}. If done with respect to $x_0$, we call it \textcolor{blue}{$x_0$-space} and use \textcolor{blue}{blue}. We will also add ``through the denoiser'' if the \textcolor{red}{$x_t$-space} projection is done using $c(d_\theta(x_t))$ and ``renoise'' if any noise is added after the projection.

\begin{wraptable}[18]{r}{0.48\textwidth}
\vspace{-1.3em}
\centering
\caption{Summary of related works, showing in which space they operate, if they reapply the kernel, and whether they propagate information through $d_\theta$.}
\label{tab:algo_mapping}

\setlength{\tabcolsep}{3pt}
\renewcommand{\arraystretch}{0.95}

\resizebox{\linewidth}{!}{%
\begin{tabular}{lccc}
\toprule
Method & Space & \shortstack{Kernel\\reapplications} & Prop. $d_\theta$ \\
\midrule
\multicolumn{4}{l}{\textit{Constraints as observations}} \\
\hspace{1mm} DPS \cite{chung2024diffusionposteriorsamplinggeneral} 
& \textcolor{red}{$x_t$} & \xmark & \cmark \\
\hspace{1mm} MMPS \cite{mmps} 
& \textcolor{red}{$x_t$} & \xmark & \cmark \\
\hspace{1mm} ReSample/HDC \cite{song2024solvinginverseproblemslatent} 
& \textcolor{blue}{$x_0$} & \cmark & \xmark \\

\midrule
\multicolumn{4}{l}{\textit{Constrained sampling}} \\
\hspace{1mm} CPS \cite{narasimhan2025constrainedposteriorsamplingtime} 
& \textcolor{blue}{$x_0$} & \xmark & \xmark \\
\hspace{1mm} \citet{zampini2025trainingfreeconstrainedgenerationstable} 
& \textcolor{red}{$x_t$} & \xmark & \xmark \\
\hspace{1mm} TS \cite{huang2024constraineddiffusiontrustsampling} 
& \textcolor{red}{$x_t$} & \cmark & \cmark \\
\hspace{1mm} PDM \cite{christopher2024constrainedsynthesisprojecteddiffusion} 
& \textcolor{red}{$x_t$} & \xmark & \xmark \\
\hspace{1mm} CDIM \cite{jayaram2025linearlyconstraineddiffusionimplicit} 
& \textcolor{red}{$x_t$} & \xmark & \xmark \\

\hspace{1mm} \textbf{PPR (Ours)} 
& \textcolor{red}{$x_t$} & \cmark & \cmark \\
\bottomrule
\end{tabular}%
}
\end{wraptable}

\paragraph{Projection-based constrained sampling.}
A number of methods enforce constraints by modifying reverse-time sampling with projections. 
\citet{christopher2024constrainedsynthesisprojecteddiffusion} (\textcolor{red}{$x_t$-space}) and \citet{liu2024mirrordiffusionmodelsconstrained} focus on convex constraints, while \citet{jayaram2024constraineddiffusionimplicitmodels, jayaram2025linearlyconstraineddiffusionimplicit} study linear constraints. Both are \textcolor{red}{$x_t$-space} and the latter project through the denoiser.

\mbox{}\citet{zampini2025trainingfreeconstrainedgenerationstable} (\textcolor{red}{$x_t$-space}) use proximal Langevin dynamics within the latent $x_t$ space, propagating gradients through the decoder. Their approach has the side effect of mitigating potential deviations from $\mathcal{S}$ by leveraging a structured latent space where the decoder can correct samples. In contrast, \citet{narasimhan2025constrainedposteriorsamplingtime} (\textcolor{blue}{$x_0$-space}) operate solely in $x_0$-space, which can cause samples to be inconsistent with the data distribution, i.e. $x_0 \notin \mathcal{S}$. For example, discontinuities can appear in continuous data (see, in their appendix, Figures 6 and 10). Adding more information via constraints can mask this issue. \citet{huang2024constraineddiffusiontrustsampling} (\textcolor{red}{$x_t$-space} through the denoiser with renoising) propose a heuristic trust region method based on the magnitude of the predicted noise $\epsilon_\theta \propto x_t - d_\theta(x_t)$. We refer to this method as TS for Trust Sampling. \citet{christopher2026constraineddiffusionproteindesign} (\textcolor{blue}{$x_0$-space}) uses domain knowledge, global constraints, and the forward kernel for protein design. In the flow matching context, \citet{eci} (\textcolor{blue}{$x_0$-space}) use extrapolation, a known exact correction, and interpolation to generate the constrained samples.

\paragraph{Posterior sampling under observation processes.}
A distinct yet related body of work focuses on sampling from posteriors derived from observation processes. While constrained sampling enforces strict adherence to a feasible set, observation-based methods typically condition generation on external measurements, often assuming an underlying Gaussian likelihood $p(y | x) = \mathcal{N}(y \mid A(x), \sigma^2)$, where $A$ is the observation process and $y$ is the observation value, as proposed in \citep{meng2024diffusionmodelbasedposterior, song2023pseudoinverseguided, chung2024diffusionposteriorsamplinggeneral, mmps, galashov2025learnguidediffusionmodel}. Theoretically, hard constraints can be framed as a special case of inverse problems by modeling the observation likelihood as a Dirac delta distribution $\mathcal{N}(y \mid A(x), \sigma^2) \xrightarrow{\sigma^2 \to 0} \delta(y - A(x)) $. However, applying standard posterior sampling algorithms \citep{daras2024surveydiffusionmodelsinverse, ye2024tfgunifiedtrainingfreeguidance} in this regime presents challenges, as these methods are generally designed for more informative likelihoods rather than the narrow distributions required for strict constraint satisfaction. Notably among these, \citet{song2024solvinginverseproblemslatent} (\textcolor{blue}{$x_0$-space}) focus on constraints and do renoising with a hand-designed kernel. \citet{huang2024DiffusionPDE} (DiffusionPDE) guide an unconditional prior with observation and PDE-residual gradients. \citet{yao2025Guided} (FunDPS) train a function-space neural-operator diffusion prior and then apply likelihood-gradient guidance.

\section{Experiments} \label{sec:experiment}

\begin{figure}
    \centering
    \includegraphics[width=0.95\linewidth]{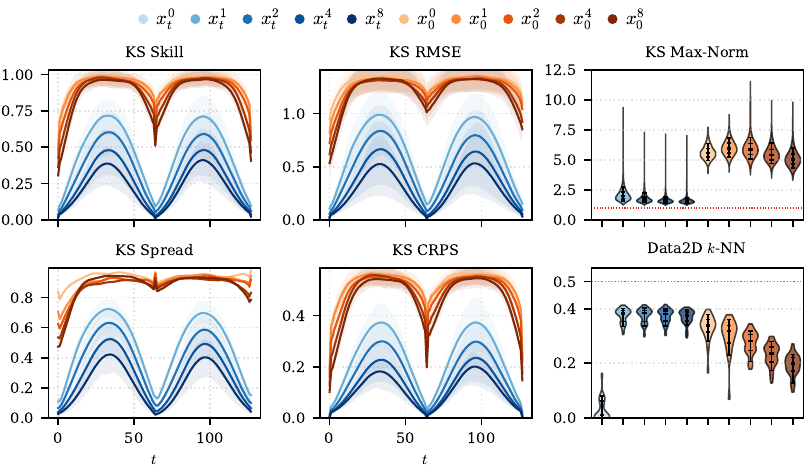}
    \vspace{-0.1cm}
    \caption{\textsc{KS} and \textsc{Data2D} PPR ablation. We vary the number of kernel applications $M$ and whether information is propagated through the denoiser. Projecting in $x_0$-space rather than $x_t$-space worsens the constrained distribution for both datasets and in \textsc{KS}, $x_0$-space samples are not in $\mathcal{S}$ because they are discontinuous. When projecting through the denoiser in $x_t$-space, larger $M$ improves the constrained distributions in both settings. For \textsc{KS}, skill, RMSE, and CRPS measure prediction accuracy; for \textsc{Data2D}, $k$-NN measures constrained-distribution quality. Max-Norm is better when closer to $1$ (dotted red line). In the \textsc{KS} metrics, the x-axis $t$ denotes system time, not diffusion time.}
    \label{fig:ablation-metrics}
    \vspace{-0.3cm}
\end{figure}

\paragraph{Protocol.} We consider three experimental settings: \textsc{Data2D}, Kuramoto-Sivashinsky (\textsc{KS}), and \textsc{Weather}. \textsc{Data2D} comprises 2D problems where the ground truth constrained distribution can be approximated. \textsc{KS} is based on the 1D Kuramoto-Sivashinsky equation \citep{kuramoto} and assesses our method and the baselines in a high-dimensional setting. Finally, \textsc{Weather} uses Appa, a global latent weather model \citep{appa}, to test PPR in a large-scale, realistic setting. The full numerical results, additional experiments, and qualitative samples are shown in the Appendices~\ref{app:detailed-results}, \ref{app:abl}, \ref{app:ks-additional}, and~\ref{app:samples}. The code is available at \url{https://github.com/OmerRochman/ppr}. Numerically, we consider $x \in \mathcal{C}$ if $c(x) \leq 4 \times 10^{-4}$ for \textsc{KS} and $c(x) \leq 4 \times 10^{-6}$ for \textsc{Data2D}.

\paragraph{\textsc{Data2D}.} We first train diffusion models on two distinct 2D distributions, a uniform checkerboard (Fig.~\ref{fig:toy-clouds}, top), and a mixture of Gaussians, which serve as priors. The term ``prior'' refers to these learned distributions $p(x_0)$, not to the terminal distribution $p_T$, which remains Gaussian. We then generate $12$ random functions $f(x)$ by sampling cosine series. For each function, we define the constraint $c(x) = 1 - e^{-f(x)^2}$, so that $\mathcal{C}$ is the zero-level set of $f$, and approximate the ground-truth (GT) constrained distribution by rejection sampling from the diffusion model prior with a small tolerance $\epsilon$, that is, $\mathrm{GT} = \{x \in \mathcal{S} \,:\, c(x) \le \epsilon\}$. The accepted samples are then refined with a few deterministic gradient steps to better satisfy the constraint. This results in $12$ constraints for each of the $2$ distributions, giving $24$ total experimental settings. For each of these, we generated ensembles with PPR and the baselines to approximate the GT constrained distribution. More details can be found in Appendix~\ref{app:data}. To assess the quality of the resulting constrained distributions, we use the $k$-nearest-neighbor ($k$-NN) cross-edge rate \citep{schilling1986knn}. The $k$-NN cross-edge rate constructs a graph on the union of GT and model samples, and counts the number of edges that connect a GT sample to a model sample. The cross-edge rate (cross edges divided by total edges) is expected to be $0.5$ when the two distributions match. The results are averaged across all prior distributions and all $12$ constraints.

\paragraph{\textsc{Kuramoto-Sivashinsky}.} We train a diffusion prior to generate trajectories of the 1D \textsc{KS} equation $u_t = - u_{xx} - u_{xxxx} - uu_x$. To avoid confusion with diffusion time, we denote the dynamical system time with a superscript and generated samples with a subscript $u_0$, which is omitted for the ground truth. We generate a dataset by sampling cosine series initial conditions $u^0 \in \mathbb{R}^{512}$ and solving the equation with periodic boundary conditions on $x \in [0, L]$, resulting in trajectories $u^{1:1024}$. We then select a subset of each trajectory $u^{512:1024} \in \mathbb{R}^{512 \times 512}$.  This is done to ensure that the system has reached the statistically stationary regime. To form the dataset, we downsize each trajectory to $\mathbb{R}^{128 \times 128}$. As evaluating the quality of approximations of conditional distributions for high-dimensional systems is an open problem \citep{linhart2023lc2st}, we rely on proxy quantities. To test the distribution of samples, we frame the perfect nonlinear observation $c(u_0) = \sum_i \| \sin(u_0^i) - \sin(u^i)  \|^2$, $i \in \{0, 64, 127 \}$ as a constraint. This is not a PDE-forecasting task but constrained trajectory generation, or perfect nonlinear data assimilation. In Appendix~\ref{app:ks-additional}, we test other constraints, including conservation of the spatial mean. We observe each of the $256$ test trajectories and for each method we generate an ensemble of $32$ trajectories. The system is deterministic, so we expect to observe close agreement with the trajectory the observation was obtained from, i.e., a delta distribution. To judge the quality of the constrained distribution, we present $4$ metrics that are best considered together. The root-mean-squared-error (RMSE) measures how far members of the ensemble are from the ground truth, on average. The skill measures how far the ensemble mean is from the ground truth, that is, the bias of the ensemble. The spread measures the deviation of the ensemble from said mean, i.e., how close it is to the delta. Lastly, the continuous ranked probability score (CRPS) penalizes both bias and misrepresentation of uncertainty (too much or too little spread). See Appendix~\ref{app:metrics} for their definitions.

\paragraph{PPR ablation.} 
Proposition \ref{prop:station} indicates that renoising and repeated kernel application are fundamental requirements in approximating the constrained marginals and we argue that using the denoiser in the projection loop has key benefits. We justify the claims on \textsc{KS} and \textsc{Data2D} by carefully ablating each component and showing that without any one of them performance degrades dramatically. To that end, we introduce a PPR variant that does not use the denoiser in the projection. In the variant, labeled $x_0^M$ in the figures, we project $x_0$ directly ($x_0$-space), without the denoiser, which is only used to get the initial $x_0 = d_\theta(x_t)$. That makes the projection $\Pi(x_0) = \arg\min_x \, c(x)$ and the transition kernel $K_t^{\text{var}}(x_t'|x_t) = p\big(x_t' |  \Pi(x_0)\big)$. For PPR and the variant, we study no renoising ($M=0$), only renoising ($M=1$) and renoising plus reapplication of the $K_{t}^{\mathrm{PPR}}$ kernel ($M>1$).

\begin{figure}
\centering

\begin{subfigure}[t]{0.49\textwidth}
  \centering
  \includegraphics[width=0.95\linewidth]{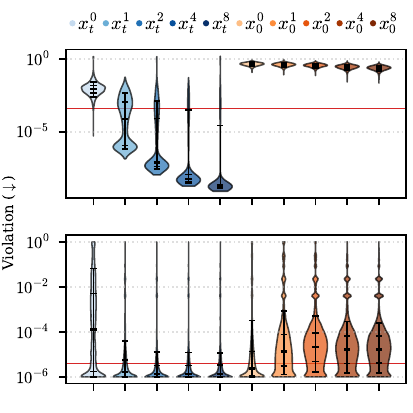}
  \caption{We vary the number of kernel applications $M$ ($x^M$) and whether information is propagated through the denoiser ($x_t^M$), or not ($x_0^M$). Projecting in $x_0$-space instead of $x_t$-space increases constraint violation; on \textsc{KS}, no $x_0$-space sample lies in $\mathcal{C}$. When projecting through the denoiser in $x_t$-space, larger $M$ lowers violation.}
  \label{fig:ablation-viol}
\end{subfigure}
\hfill
\begin{subfigure}[t]{0.49\textwidth}
  \centering
  \includegraphics[width=0.95\linewidth]{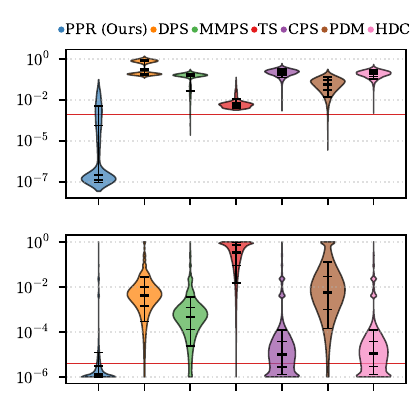}
  \caption{PPR obtains the lowest constraint violation and highest feasibility $\%$. The $\%$ of feasible samples is over $82\%$ for PPR and $8\%$ for the next best method on \textsc{KS} and $79\%$ vs $31\%$ on \textsc{Data2D}.}
  \label{fig:baseline-viol}
\end{subfigure}

\caption{Constraint violation for \textsc{KS} (top) and \textsc{Data2D} (bottom) for the ablation (left) and the baselines comparison (right). Red lines show the thresholds, $4 \times 10^{-4}$ for \textsc{KS} and $4 \times 10^{-6}$ for \textsc{Data2D}. Samples above are considered not in $\mathcal{C}$.}
\label{fig:constraint-violations}
\vspace{-0.3cm}

\end{figure}

We run the ablations on \textsc{KS} and \textsc{Data2D}, confirming the necessity of the projection through the denoiser, renoising, and multiple kernel applications. Figures~\ref{fig:ablation-metrics} and~\ref{fig:ablation-viol}  show that not using the denoiser (in orange) results in samples that score worse on all metrics and have much higher constraint violation ($x_0 \notin \mathcal{C}$), with the number of kernel applications doing little to improve the situation. The Max-Norm shows that the \textsc{KS} samples are not in $\mathcal{S}$ as they present sharp discontinuities. The advantage of the denoiser (in blue) is clear, the samples remain in $\mathcal{S}$ while improving all metrics and the constraint violation, achieving $x_0 \in \mathcal{C}$. Increasing the number of kernel applications $M$ consistently improves the samples at a linear computational cost. No renoising, i.e. $M=0$ is omitted from the \textsc{KS} subfigures as it diverged, supporting the necessity of $M>0$. These conclusions are confirmed on \textsc{Data2D}. $x_t$-space variants achieve $k$-NN cross-edge rates closer to $0.5$ and lower constraint violation. $M=0$ fails catastrophically but the constraint violation improves as $M$ increases. On this low dimensional distribution, increasing $M>1$ quickly saturates (see Appendix~\ref{app:thm}), resulting in diminishing returns that do not appear in the higher dimensional settings.

\begin{figure}
    \centering
    \includegraphics[width=0.95\linewidth]{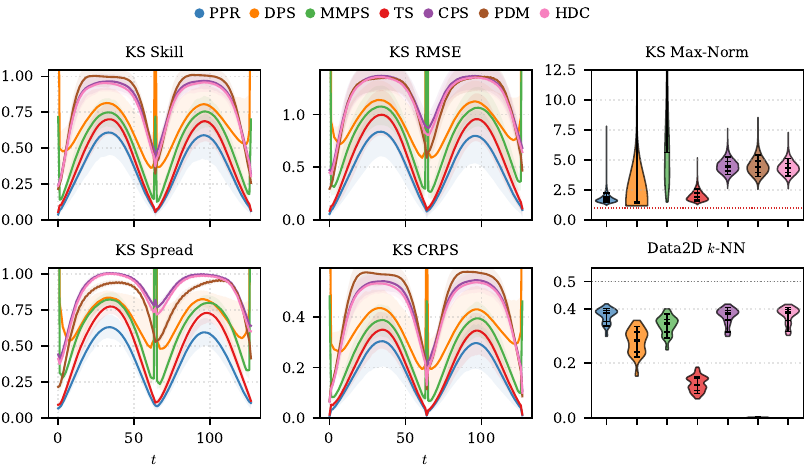}
    \vspace{-0.1cm}
    \caption{Baseline comparison on \textsc{KS} and \textsc{Data2D} (bottom right). PPR gives the best constrained distribution in both cases. Only PPR and TS produce samples in $\mathcal{S}$; all other methods yield discontinuous samples. For \textsc{KS}, skill, RMSE, and CRPS measure prediction accuracy; for \textsc{Data2D}, the $k$-NN cross-edge rate measures constrained-distribution quality. Max-Norm is better when closer to $1$ (dotted red line). In the \textsc{KS} metrics, the x-axis $t$ denotes system time, not diffusion time.}
    \label{fig:baseline-metrics}
    \vspace{-0.3cm}
\end{figure}

\begin{wraptable}{r}{0.55\textwidth}
\vspace{-1.3em}
\centering
\caption{Data2D comparison. The feasibility threshold is $4 \times 10^{-6}$. The $k$-NN rate is better closer to $0.5$.}
\setlength{\tabcolsep}{4pt}
\begin{tabular*}{\linewidth}{@{\extracolsep{\fill}} l r r r}
\toprule
Method & Time (s) & Feas.\,$\uparrow$ & $k$-NN \\
\midrule
PPR (Ours) & 14.9$\,\pm\,$0.6 & \textbf{79.0\%} & \textbf{0.375$\,\pm\,$0.026} \\
DPS        & 2.6$\,\pm\,$0.2  & 1.0\%           & 0.279$\,\pm\,$0.046 \\
MMPS       & 11.7$\,\pm\,$0.4 & 3.6\%           & 0.340$\,\pm\,$0.034 \\
Trust      & 3.8$\,\pm\,$0.2  & 0.1\%           & 0.124$\,\pm\,$0.027 \\
CPS        & 5.9$\,\pm\,$0.4  & 31.6\%          & 0.373$\,\pm\,$0.030 \\
PDM        & 10.8$\,\pm\,$0.1 & 1.4\%           & 0.002$\,\pm\,$0.000 \\
HDC        & 3.5$\,\pm\,$0.1  & 30.3\%          & \textbf{0.375$\,\pm\,$0.030} \\
\bottomrule
\end{tabular*}

\vspace{2pt}
\label{tab:data2d-main-text}
\vspace{-0.4cm}
\end{wraptable}

\paragraph{Baselines comparison.} We now compare PPR to other constrained sampling methods. The baselines are four explicit constrained samplers (TS, CPS, PDM, and HDC) and two general posterior samplers (DPS and MMPS), plus DiffusionPDE on \textsc{KS}. All are training-free and, within each experiment, all use the same learned model, isolating the effect of the sampler. Figures~\ref{fig:baseline-viol} and~\ref{fig:baseline-metrics}, and Tables~\ref{tab:data2d-main-text} and~\ref{tab:ks-main-text} show that PPR has the highest feasibility rate, the lowest constraint violation, and produces a closer approximation to the constrained distribution as measured by the Skill, RMSE, and the $k$-NN cross-edge rate on \textsc{Data2D} and CRPS on \textsc{KS}. PDM, CPS, and HDC, working in $x_0$-space, fail to produce $x_0 \in \mathcal{S}$ as the \textsc{KS} samples are discontinuous. The percentage of feasible samples is over $82\%$ for PPR vs $8\%$ for the next best method on \textsc{KS} and $79\%$ vs $31\%$ on \textsc{Data2D}.

Cost is a secondary objective: a sample that violates the constraint or leaves the data distribution is useless. The cost is tunable through the number of diffusion steps, project-renoise applications, and inner projection iterations. All methods used the same computational budget. However, many baselines do not offer an effective way to trade compute for improvements. For DPS and MMPS, the main lever is the number of diffusion steps, which quickly hit diminishing returns. CPS, PDM, and HDC also expose the number of iterations in their projection routines, but these $x_0$-space projections converge quickly, often to points outside $\mathcal{S}$, so additional iterations bring little benefit. In TS, the trust-region mechanism can terminate projection prematurely and even with a large region its normalized projection is less effective than the off-the-shelf solvers used by PPR, leading to wasted computation.

\begin{table}[t]
\centering
\caption{Comparison on \textsc{KS}. Only PPR produces feasible samples $x\in\mathcal{C}\cap\mathcal{S}$ at a meaningful rate. The feasibility threshold was set at $c(x) < 4\times 10^{-4}$. Max-Norm is better the closer it is to $1$.}
\label{tab:ks-main-text}

\resizebox{\linewidth}{!}{%
\begin{tabular}{l | c c c c c c c}
\toprule
Method & Feas.\,(\%) $\uparrow$ & RMSE $\downarrow$ & CRPS $\downarrow$ & Skill $\downarrow$ & Spread $\downarrow$ & Max-Norm & Time\,(s) $\downarrow$ \\
\midrule
    PPR (Ours) & {\boldmath$82.7$} & {\boldmath$0.503$} & {\boldmath$0.177$} & {\boldmath$0.365$} & {\boldmath$0.385$} & {\boldmath$1.80$} & $80.2$ \\
    DPS & $0.0$ & $1.089$ & $0.352$ & $0.666$ & $0.865$ & $17.11$ & $25.5$ \\
    MMPS & $1.0$ & $0.830$ & $0.265$ & $0.519$ & $0.672$ & $19.28$ & $307$ \\
    TS & $0.0$ & $0.614$ & $0.206$ & $0.427$ & $0.487$ & $2.01$ & $398$ \\
    CPS & $0.0$ & $1.179$ & $0.436$ & $0.805$ & $0.890$ & $4.50$ & $50.8$ \\
    PDM & $0.8$ & $1.137$ & $0.463$ & $0.831$ & $0.798$ & $4.48$ & $296$ \\
    HDC & $0.0$ & $1.152$ & $0.425$ & $0.787$ & $0.870$ & $4.36$ & $49.8$ \\
    DiffusionPDE & $8.2$ & $1.035$ & $0.374$ & $0.707$ & $0.788$ & $2.80$ & $91.5$ \\
\bottomrule
\end{tabular}%
}
\end{table}

\begin{figure}[!t]
    \centering
    \vspace{-0.3cm}
    \includegraphics[width=0.95\linewidth]{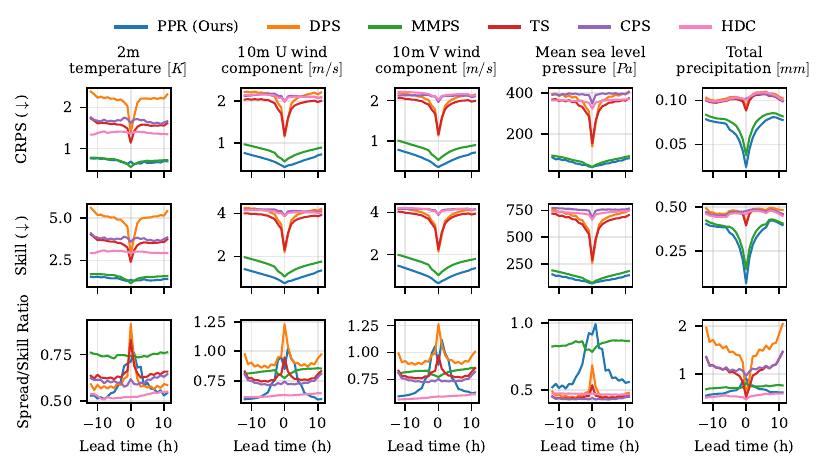}
    \vspace{-0.3cm}
    \caption{Comparison on Appa across $4$ test dates. PPR outperforms all baselines on all variables. ERA5 surface variables shown. The metrics were computed with WeatherBench2 \citep{rasp2024weatherbench2benchmarkgeneration}.}
    \label{fig:appa}
    \vspace{-0.3cm}
\end{figure}

\paragraph{Large-scale latent weather.} In a setting similar to \textsc{KS}, we treat a perfect nonlinear observation as a constraint. We use Appa \citep{appa}, a score-based model that generates global atmospheric trajectories at $0.25^\circ$ resolution and $1$-hour intervals trained on ERA5 \citep{era5}. One atmospheric state has shape $1440 \times 721 \times 71$, which is $\mathcal{O}(10^8)$ elements $\approx 300$MB. The decoder $\mathcal{D}$, through which we define the constraint, has $170$M parameters. Appa's denoiser generates $24$-hour trajectories, from which we observe 12:00 PM through the decoder. This results in the constraint $c(x_0) = \| \mathcal{D}(x_0^{12}) - y  \|_2^2$, where $y$ is the value of the observation, extracted from ERA5.
Figure~\ref{fig:appa} shows the results of the constrained sampling across $4$ test dates. For each method, an ensemble of size $32$ was sampled under the same computational budget and wall runtime. The computational budget is the same as \citep{appa}, and was not optimized. PPR outperforms all baselines: it obtains lower constraint violation and its ensemble trajectories remain closer to the true trajectory for longer. 
See Appendix~\ref{app:detailed-results} and~\ref{app:samples} for more results.

\section{Conclusion}
To address the problem of constrained sampling in score-based models, we defined a constrained forward process with its corresponding reverse process, identified a transition kernel whose unique stationary distribution is that constrained marginal, and introduced PPR as an approximation of that kernel. Over a series of experiments, we showed that PPR achieves a higher feasibility rate, lower constraint violation, and closer approximation to the constrained distribution compared to the state of the art and that each of the components of PPR, the projection through the denoiser, the renoising, and the reapplication of the kernel, is fundamental to its performance. Additionally, PPR offers a meaningful way of trading compute for sample and constrained distribution quality.

\paragraph{Limitations.} 
We rely on approximations to sample $p_t^\mathcal{C}(x_0 | x_t)$ (see Appendix~\ref{app:thm} for the error bound). PPR relies on the differentiability of the constraint function $c(x)$ to propagate information from $x_0$ to $x_t$. PPR applies to differentiable constraints whose projection is numerically tractable, and does not assume a closed-form or domain-specific projection. Highly discontinuous, discrete, combinatorial, or black-box constraints are outside the current scope. To extend PPR to another domain, one needs a pretrained denoiser and a differentiable $c$ that measures violation of the desired property, for example engineering design requirements or geometric constraints, although we have not evaluated these applications. PPR also has a computational overhead, as each reverse step requires repeated projections through the denoiser and renoising. This cost grows linearly with $M$ and can be traded against feasibility by tuning $M$ and the number of projection steps.

\paragraph{Future work.} A possible way of reducing the computational cost of PPR is to only project at a subset of diffusion steps. While we preliminarily found that it gives poorer constraint satisfaction and constrained distribution approximation, it is an additional lever to be tuned by the user. Further, extending beyond equality constraints to mixed and inequality constraints is a promising direction. Inequality constraints could be handled by a projection with momentum or Langevin projection that stops once inside $\mathcal{C}$. Extending the framework to explicitly handle discrete constraints and domains such as inverse folding in constrained protein design also remains open for further research.

\FloatBarrier

\begin{ack}
This work was partially supported by ServicePublic de Wallonie Recherche under grant n° 2010235 – ARIAC by DIGITALWALLONIA4.AI. 
Computational resources have been provided by the Consortium des Équipements de Calcul Intensif (CÉCI), funded by the Fonds de la Recherche Scientifique de Belgique (F.R.S.-FNRS) under Grant No. 2.5020.11 and by the Walloon Region. 
The present research benefited from computational resources made available on Lucia, the Tier-1 supercomputer of the Walloon Region, infrastructure funded by the Walloon Region under the grant agreement n°1910247.
\end{ack}

\bibliography{example_paper}
\bibliographystyle{plainnat}

\newpage
\appendix

\section{Metrics} \label{app:metrics}

We use the same metrics as in \citet{appa}, and partially reproduce their text for the definitions here. The evaluation is performed using WeatherBench2 \cite{rasp2024weatherbench2benchmarkgeneration} for Appa data. For the \textsc{KS} experiments, the metrics are the same, but with one spatial dimension removed.

\paragraph{Skill.}
Skill is computed as the root mean square error of the posterior mean of an ensemble compared to the ground-truth trajectory. For $K$ ensembles each consisting of $M$ predicted states $\hat{x}$ of resolution $H \times W$, ground truth $x$, the skill of a single time step is computed as
\[
    \text{Skill} = \sqrt{\dfrac{1}{KHW} \sum_{k=1}^{K} \sum_{i=1}^{H} \sum_{j=1}^{W} \left(x_{i,j}^k - \dfrac{1}{M}\sum_{m=1}^M \hat{x}_{i,j}^{k,m}\right)^2}.
\]

\paragraph{Spread.}
Ensemble spread is computed as the square root of the ensemble variance \cite{fortin2014should}:
\[
    \text{Spread} = \sqrt{\dfrac{1}{KHW} \sum_{k=1}^{K} \sum_{i=1}^{H} \sum_{j=1}^{W} \dfrac{1}{M-1} \sum_{m=1}^{M}\left(\hat{x}_{i,j}^{k,m} - \dfrac{1}{M}\sum_{n=1}^M \hat{x}_{i,j}^{k,n}\right)^2}.
\]

\paragraph{Spread-skill ratio.}
A well-calibrated forecast should have a (corrected for ensemble size) spread-skill ratio of 1, which is a necessary but not sufficient condition. Ratios below one indicate overconfident estimations. The correct ratio is defined as
\[
    \text{Ratio} = \sqrt{\dfrac{M+1}{M}} \dfrac{\text{Spread}}{\text{Skill}}.
\]

\paragraph{Continuous ranked probability score (CRPS).}
The CRPS \cite{crps} is defined as
\[
    \text{CRPS} = \dfrac{1}{K} \sum_{k=1}^K \left(\dfrac{1}{M}\sum_{m=1}^M || \hat{x}^{k,m} - x^k ||_{L_1} - \dfrac{1}{2M(M-1)} \sum_{m=1}^M\sum_{n=1}^{M} || \hat{x}^{k,m} - \hat{x}^{k,n} ||_{L_1}\right).
\]
The first term penalizes the average divergence from the ground truth, while the second term encourages spread. Therefore, the CRPS is lowest when the distribution of the ensemble matches the ground-truth distribution.

\FloatBarrier

\section{Data} \label{app:data}

\subsection{\textsc{Data2D}}

We generated i.i.d.\ samples in $\mathbb{R}^2$ from two parametric distributions:  a uniform checkerboard distribution and a ``banana'' Gaussian mixture. Unless otherwise stated, reported samples are standardized by using an empirically estimated mean $\mu\in\mathbb{R}^2$ and standard deviation $\sigma\in\mathbb{R}^2$, i.e.,
\[
\tilde{x} \;=\; \frac{(x-\mu)}{\sigma}.
\]
The normalization statistics are estimated from Monte Carlo samples drawn from the \emph{non-standardized} distribution.

\paragraph{Checkerboard.}
Fix an integer grid size $m$ and a jitter scale $\eta\ge 0$. We draw $i\sim \mathrm{Unif}\{0,\dots,m-1\}$ and then draw $j$ uniformly from $\{0,\dots,m-1\}$ subject to the parity constraint $(i+j)\bmod 2 = 0$ (equivalently, points occupy alternating squares). Given $(i,j)$, we draw $u_1,u_2\sim\mathcal U(0,1)$ independently and set
\[
x \;=\; \begin{bmatrix} i+u_1 \\ j+u_2 \end{bmatrix}.
\]
If $\eta>0$, we add isotropic Gaussian jitter, $x \leftarrow x + \eta\,\xi$ with $\xi\sim\mathcal N(0,I_2)$.

\paragraph{Banana Gaussian mixture model (Banana-GMM).}
Let $\{w_k\}_{k=1}^K$ be nonnegative mixture weights with $\sum_k w_k=1$, and let $\mu_k\in\mathbb{R}^2$ and $s_k\in\mathbb{R}_+^2$ be component means and diagonal standard deviations. We first sample a component index $k\sim\mathrm{Categorical}(w_1,\dots,w_K)$ and then draw a Gaussian sample
\[
z \sim \mathcal N\!\bigl(\mu_k,\;\mathrm{diag}(s_k^2)\bigr), \qquad z=\begin{bmatrix} z_1 \\ z_2 \end{bmatrix}.
\]
We then apply a ``banana'' shear in the second coordinate,
\[
x_1 = z_1,\qquad
x_2 = z_2 + b\bigl(z_1^2 - \mathbb{E}[z_1^2]\bigr),
\]
where $b\in\mathbb{R}$ controls the curvature and the centering term $\mathbb{E}[z_1^2]$ is computed analytically under the mixture,
\[
\mathbb{E}[z_1^2] \;=\; \sum_{k=1}^K w_k\bigl((\mu_k)_1^2 + (s_k)_1^2\bigr).
\]

\paragraph{GRF Constraints.} We generated a collection of $12$ independent scalar constraint functions on $\mathbb{R}^2$ as random draws from an approximate Gaussian Random Field (GRF) \citep{grf} with an RBF kernel, using random Fourier features. All constraints are defined by the same hyperparameters. Fix the number of features $M$ (here $M=64$), lengthscale $\ell>0$ (here $\ell=0.25$ in each input dimension), kernel variance $\sigma_f^2$ (here $\sigma_f^2=1$), and an optional additive bias. For each constraint, we sample frequencies and phases
\[
\omega_m \sim \mathcal N\!\left(0,\ell^{-2} I\right),\qquad
\phi_m \sim \mathcal U(0,2\pi),
\qquad m=1,\dots,M,
\]
independently, and we sample coefficients
\[
a_m \sim \mathcal N(0,1),\qquad m=1,\dots,M.
\]
We define the random feature map for $x\in\mathbb{R}^2$ as
\[
\varphi_m(x) \;=\; \cos(\omega_m^\top x + \phi_m),
\]
and the corresponding random function
\[
f(x) \;=\; \sqrt{\frac{2\sigma_f^2}{M}}\sum_{m=1}^M a_m\,\varphi_m(x) \;+\; \beta,
\]
where the bias term is sampled as
\[
\beta \sim \mathcal N\!\left(0,\;\sigma_b^2\,\sigma_f^2\right),
\]
with $\sigma_b=0.05$. In the limit $M\to\infty$, this construction converges (in distribution) to a zero-mean GP with an RBF kernel of lengthscale $\ell$ and marginal variance $\sigma_f^2$. Each constraint is obtained by smoothing $f(x)$,
\[
c(x) \;=\; 1 -\,e^{-f(x)^2}.
\]
We formed the $12$ constraints $\{c_i\}_{i=1}^{12}$ by sampling $(\{\omega_m,\phi_m,a_m\}_{m=1}^M,\beta)$ independently for each $i$. We used a tolerance $\epsilon = 10^{-2}$ for the creation of the constrained distributions.

\subsection{\textsc{Kuramoto-Sivashinsky}}

The 1-D Kuramoto--Sivashinsky Eq.~\citep{kuramoto, sivashinksy}
\[
\partial_t u + \partial_{xx} u + \partial_{xxxx} u + u\,\partial_x u = 0
\]
is solved on the periodic domain $x\in[0,64]$ with a pseudospectral solver. The spatial derivatives use the FFT on the grid of size $N = 512$ and no de-aliasing. Time stepping was done using the first-order backward-difference formula (BDF1) with fixed step $\mathrm dt=10^{-1}$ for $1024$ steps, giving $t_{\max}=102.4$. During training, we used a subset of each solution, starting at step $512$ and ending at step $1024$, corresponding to $t \in [51.2, 102.4]$. Each implicit step is solved by Newton iterations. Initial data are random sums of ten cosine modes,
\[
u_0(x)=\sum_{k=1}^{10} a_k \cos\bigl(2 \pi\omega_k x/64+\phi_k\bigr),
\]
where amplitudes $a_k\sim\mathcal U(0,1)$, frequencies $\omega_k \sim \mathcal{U}(1, 5)$ and phases $\phi_k\sim\mathcal U(0,2\pi)$ are sampled independently. The dataset comprises $65{,}536$ training and $128$ validation trajectories, each generated with an independent initial condition. The test trajectories for Section~\ref{sec:experiment} were generated with the unconditional diffusion model.

\subsection{Architecture}
Both architectures use SiLU activations, sinusoidal noise-level embeddings, and
EDM-style preconditioning. 

\textbf{Data2D} uses a Modulated MLP: a 3-hidden-layer MLP (256 units per
layer) with residual connections.  The noise level $\sigma$ and the input
coordinates $x\in \mathbb{R}^2$ and their norm are each encoded by
separate sinusoidal embeddings and injected into every hidden layer via
FiLM modulation.

\textbf{Kuramoto-Sivashinsky (KS)} uses a UNet with channel progression
$[32, 64, 128, 256]$.  Each scale has one FiLM-modulated residual block
($3{\times}3$ convolutions) and a strided convolution (encoder) or transposed
convolution (decoder) for downsampling and upsampling.
A small MLP lifts the 1-channel input to the base feature width and projects back
at the output.  Circular padding is applied in the spatial direction; asymmetric
zero-padding handles the (non-periodic) time dimension.

The noise schedule is given by

\begin{equation}
\sigma_t=\exp\left(s\,\operatorname{logit}\left(t(1-2\varepsilon)+\varepsilon\right)+\log\sigma_{\mathrm{med}}\right),
\end{equation}

\begin{equation}
\varepsilon=\left(\sqrt{\frac{\sigma_{\min}}{\sigma_{\max}}}\right)^{1/\alpha},
\quad
\sigma_{\mathrm{med}}=\sqrt{\sigma_{\min}\sigma_{\max}},
\end{equation}

and a constant $\alpha_t=1$ scale schedule. This schedule was inspired by \citet{appa}.

\input{tables/arch_table}

\subsection{Training}

\textsc{Data2D} and \textsc{KS} were trained for 300 epochs using the Muon optimizer \citep{muon} with no weight decay and $k=1$ optimizer step per update. \textsc{Data2D} used a batch size of 1024, a peak learning rate of $10^{-3}$, no warmup, cosine decay over $101 \times 5000$ steps, and a final learning rate of $10^{-6}$. \textsc{KS} used a batch size of 64, a peak learning rate of $10^{-4}$, 100 warmup steps, cosine decay over $100{,}000$ steps, and the same final learning rate of $10^{-6}$. All models were trained on a single NVIDIA A5000-24GB. The \textsc{KS} model trained overnight while the \textsc{Data2D} model trained for $4$ h.

\subsection{Appa}

Appa was trained on ERA5 \citep{era5}. We refer the reader to \citet{appa} for details, including training and compute.

\FloatBarrier

\section{Theoretical analysis and approximation error bound} \label{app:thm}

In this section we restate Proposition~\ref{prop:station} and prove it. We also bound the error of the PPR approximation, and justify the assumptions in that bound. Both propositions assume bounded second moment for all distributions, a nondegenerate constrained set \(0<\mathbb P(x_0\in\mathcal C)<1\), and well-defined conditional constrained posteriors \(\mathbb P(x_0\in\mathcal C\mid x_t)>0\) on the relevant support.

\begin{proposition}
Let $K_t(x_t' | x_t) = \int_\mathcal{C} p(x_t'|x_0) \, p^\mathcal{C}(x_0|x_t) \, dx_0$ with $p^\mathcal{C}(x_0|x_t) = \frac{p(x_0|x_t) \, \Id(x_0)}{\mathbb{P}(x_0 \in \mathcal{C}|x_t)}.$ This kernel connects the unconstrained $p_t(x_t)$ and constrained $p_t^\mathcal{C}(x_t)$ priors at time $t$. Under this kernel, $p_t^\mathcal{C}$ is the unique stationary distribution and $(K_t)^M q \to p_t^\mathcal{C}$ as $M \to \infty$ for any initial distribution $q$.
\end{proposition}

\paragraph{Sketch.}
The kernel \(K_t\) can be viewed as a two-step resampling procedure: starting from the current noisy variable \(x_t\), it first samples a clean point \(x_0\) from the posterior restricted to the constraint set \(\mathcal C\), and then renoises this \(x_0\) to obtain a new \(x_t'\). If the current \(x_t\) is already distributed according to the constrained noisy marginal \(p_t^\mathcal C\), then this posterior-resampling step recovers exactly the constrained clean distribution, and the renoising step maps it back to \(p_t^\mathcal C\). This shows stationarity. Since the renoising distribution is Gaussian, the chain can move between any regions of the state space with positive probability, which rules out multiple closed classes or periodic behavior. Therefore the stationary distribution is unique, and repeated applications of the kernel converge to \(p_t^\mathcal C\).

\begin{proof}
Fix $t>0$ and write
\[
p_t^\mathcal{C}(x_t)
=
\frac{1}{Z^\mathcal{C}}
\int_\mathcal{C} p(x_t|x_0)p(x_0)\,dx_0,
\qquad
Z^\mathcal{C}
=
\mathbb{P}(x_0\in\mathcal{C}).
\]
We first show that $p_t^\mathcal{C}$ is stationary for $K_t$, namely
\[
\int K_t(x_t'|x_t)\,p_t^\mathcal{C}(x_t)\,dx_t
=
p_t^\mathcal{C}(x_t').
\]

By Bayes' rule,
\[
p(x_0|x_t)
=
\frac{p(x_t|x_0)p(x_0)}{p_t(x_t)}.
\]
Moreover,
\[
\mathbb{P}(x_0\in\mathcal{C}|x_t)
=
\int_\mathcal{C} p(x_0|x_t)\,dx_0
=
\frac{\int_\mathcal{C} p(x_t|x_0)p(x_0)\,dx_0}{p_t(x_t)}
=
\frac{Z^\mathcal{C}p_t^\mathcal{C}(x_t)}{p_t(x_t)}.
\]
Hence, for $x_0\in\mathcal{C}$,
\[
p^\mathcal{C}(x_0|x_t)
=
\frac{p(x_0|x_t)\Id(x_0)}
{\mathbb{P}(x_0\in\mathcal{C}|x_t)}
=
\frac{p(x_t|x_0)p(x_0)}
{Z^\mathcal{C}p_t^\mathcal{C}(x_t)}.
\]

Substituting this expression into the stationarity equation gives
\[
\begin{aligned}
\int K_t(x_t'|x_t)p_t^\mathcal{C}(x_t)\,dx_t
&=
\int
\int_\mathcal{C}
p(x_t'|x_0)
p^\mathcal{C}(x_0|x_t)
\,dx_0\,
p_t^\mathcal{C}(x_t)\,dx_t \\
&=
\int
\int_\mathcal{C}
p(x_t'|x_0)
\frac{p(x_t|x_0)p(x_0)}
{Z^\mathcal{C}p_t^\mathcal{C}(x_t)}
\,dx_0\,
p_t^\mathcal{C}(x_t)\,dx_t \\
&=
\frac{1}{Z^\mathcal{C}}
\int
\int_\mathcal{C}
p(x_t'|x_0)p(x_t|x_0)p(x_0)
\,dx_0\,dx_t \\
&=
\frac{1}{Z^\mathcal{C}}
\int_\mathcal{C}
p(x_t'|x_0)p(x_0)
\left(
\int p(x_t|x_0)\,dx_t
\right)
dx_0 \\
&=
\frac{1}{Z^\mathcal{C}}
\int_\mathcal{C}
p(x_t'|x_0)p(x_0)\,dx_0 \\
&=
p_t^\mathcal{C}(x_t').
\end{aligned}
\]
Thus $p_t^\mathcal{C}$ is stationary under $K_t$.

It remains to justify uniqueness and convergence. Since \(\sigma_t>0\), the renoising density
\[
p(x_t'\mid x_0)=\mathcal N(x_t';\alpha_t x_0,\sigma_t^2 I)
\]
is strictly positive for all \(x_t'\). Hence $K_t(x_t'\mid x_t) > 0$, for all \(x_t,x_t'\). Therefore \(K_t\) is Lebesgue-irreducible and aperiodic. Since \(p_t^\mathcal C\) is an invariant probability measure, standard general-state-space Markov chain theory \citep{MC_book} implies that \(K_t\) is positive Harris recurrent, that \(p_t^\mathcal C\) is the unique stationary distribution, and that for every initial distribution \(q\),
\[
(K_t)^M q \to p_t^\mathcal{C} \qquad \text{as } M\to\infty
\] 
for any initial distribution $q$ in the total variation sense. This proves the claim.
\end{proof}

\begin{proposition} \label{th:error}
Assume $x_t \mapsto d_\theta(\Pi_d(x_t))$ is $L_t$-Lipschitz and that $\alpha_t L_t < 1$. Then
\[
W_2\!\left(q_t^{(M)}, p_t^\mathcal{C}\right)
\leq
(\alpha_t L_t)^M \, W_2\!\left(q_t^{(0)}, \tilde{p}_t^\mathcal{C}\right)
+
\frac{\Delta_t}{1 - \alpha_t L_t},
\]
where
\[
\Delta_t
=
\sup_{x_t}
W_2\!\left(
K_t^{\mathrm{PPR}}(\,\cdot \mid x_t),
K_t^*(\,\cdot \mid x_t)
\right).
\]
\end{proposition}

\paragraph{Sketch.}
The error is split into two effects: the finite number of PPR iterations and the fact that the PPR kernel is only an approximation of the ideal constrained kernel. The first effect is controlled by showing that the PPR update is contractive: two nearby noisy inputs are mapped to Gaussian transitions whose means remain closer by a factor at most \(\alpha_t L_t<1\). Hence repeated PPR steps forget the initialization geometrically fast. The second effect is the limiting bias caused by replacing the ideal kernel with the approximate one. This bias is controlled by the largest one-step discrepancy \(\Delta_t\) between the two kernels. Because the approximate kernel is contractive, these one-step errors do not accumulate without bound; instead they are amplified by at most the geometric factor \(1/(1-\alpha_t L_t)\). Combining the geometric mixing error with this accumulated approximation bias gives the stated bound.

\begin{proof}
Let
\[
\rho_t := \alpha_t L_t .
\]
By assumption, $\rho_t<1$. We decompose
\[
W_2\!\left(q_t^{(M)},p_t^\mathcal{C}\right)
\leq
W_2\!\left(q_t^{(M)},\tilde{p}_t^\mathcal{C}\right)
+
W_2\!\left(\tilde{p}_t^\mathcal{C},p_t^\mathcal{C}\right).
\]

We first bound the mixing term. The PPR kernel is
\[
K_t^{\mathrm{PPR}}(x_t'|x_t)
=
\mathcal{N}\!\left(
x_t';
\alpha_t d_\theta(\Pi_d(x_t)),
\sigma_t^2 I
\right).
\]
For any $x_t,\bar{x}_t$, the two transition distributions have the same covariance. Hence
\[
\begin{aligned}
W_2\!\left(
K_t^{\mathrm{PPR}}(\cdot|x_t),
K_t^{\mathrm{PPR}}(\cdot|\bar{x}_t)
\right)
&=
\alpha_t
\left\|
d_\theta(\Pi_d(x_t))
-
d_\theta(\Pi_d(\bar{x}_t))
\right\| \\
&\leq
\alpha_t L_t \|x_t-\bar{x}_t\| \\
&=
\rho_t \|x_t-\bar{x}_t\|.
\end{aligned}
\]
Therefore, for any probability measures $q_1,q_2$,
\[
W_2\!\left(
K_t^{\mathrm{PPR}} q_1,
K_t^{\mathrm{PPR}} q_2
\right)
\leq
\rho_t W_2(q_1,q_2).
\]
That is, $K_t^{\mathrm{PPR}}$ is a contraction.

Define
\[
q_t^{(M)}
=
\left(K_t^{\mathrm{PPR}}\right)^M q_t^{(0)}
\]
and let $\tilde{p}_t^\mathcal{C}$ be stationary for $K_t^{\mathrm{PPR}}$. It exists since $K_t^{\mathrm{PPR}}$ is a contraction. Then repeated application gives
\[
W_2\!\left(q_t^{(M)},\tilde{p}_t^\mathcal{C}\right)
\leq
\rho_t^M
W_2\!\left(q_t^{(0)},\tilde{p}_t^\mathcal{C}\right).
\]

It remains to bound the bias term. Since $\tilde{p}_t^\mathcal{C}$ is stationary for $K_t^{\mathrm{PPR}}$ and $p_t^\mathcal{C}$ is stationary for $K_t^*$,
\[
W_2\!\left(\tilde{p}_t^\mathcal{C},p_t^\mathcal{C}\right)
=
W_2\!\left(
K_t^{\mathrm{PPR}}\tilde{p}_t^\mathcal{C},
K_t^*p_t^\mathcal{C}
\right).
\]
By the triangle inequality,
\[
\begin{aligned}
W_2\!\left(
K_t^{\mathrm{PPR}}\tilde{p}_t^\mathcal{C},
K_t^*p_t^\mathcal{C}
\right)
&\leq
W_2\!\left(
K_t^{\mathrm{PPR}}\tilde{p}_t^\mathcal{C},
K_t^{\mathrm{PPR}}p_t^\mathcal{C}
\right) \\
&\quad+
W_2\!\left(
K_t^{\mathrm{PPR}}p_t^\mathcal{C},
K_t^*p_t^\mathcal{C}
\right).
\end{aligned}
\]
The first term is bounded by the contraction estimate:
\[
W_2\!\left(
K_t^{\mathrm{PPR}}\tilde{p}_t^\mathcal{C},
K_t^{\mathrm{PPR}}p_t^\mathcal{C}
\right)
\leq
\rho_t
W_2\!\left(\tilde{p}_t^\mathcal{C},p_t^\mathcal{C}\right).
\]
For the second term, coupling the same $x_t\sim p_t^\mathcal{C}$ and then using an optimal conditional coupling of the two kernels gives
\[
\begin{aligned}
W_2\!\left(
K_t^{\mathrm{PPR}}p_t^\mathcal{C},
K_t^*p_t^\mathcal{C}
\right)
&\leq
\sup_{x_t}
W_2\!\left(
K_t^{\mathrm{PPR}}(\cdot|x_t),
K_t^*(\cdot|x_t)
\right) \\
&=
\Delta_t .
\end{aligned}
\]
Thus
\[
W_2\!\left(\tilde{p}_t^\mathcal{C},p_t^\mathcal{C}\right)
\leq
\rho_t
W_2\!\left(\tilde{p}_t^\mathcal{C},p_t^\mathcal{C}\right)
+
\Delta_t.
\]
Since $\rho_t<1$, rearranging yields
\[
W_2\!\left(\tilde{p}_t^\mathcal{C},p_t^\mathcal{C}\right)
\leq
\frac{\Delta_t}{1-\rho_t}
=
\frac{\Delta_t}{1-\alpha_t L_t}.
\]

Combining the mixing and bias bounds,
\[
\begin{aligned}
W_2\!\left(q_t^{(M)},p_t^\mathcal{C}\right)
&\leq
W_2\!\left(q_t^{(M)},\tilde{p}_t^\mathcal{C}\right)
+
W_2\!\left(\tilde{p}_t^\mathcal{C},p_t^\mathcal{C}\right) \\
&\leq
(\alpha_t L_t)^M
W_2\!\left(q_t^{(0)},\tilde{p}_t^\mathcal{C}\right)
+
\frac{\Delta_t}{1-\alpha_t L_t}.
\end{aligned}
\]
\end{proof}

\begin{figure}
    \centering
    \includegraphics[width=0.5\linewidth]{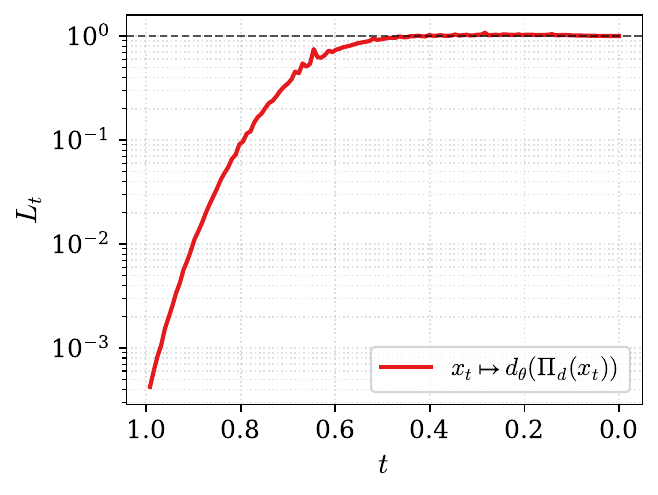}
    \caption{Lipschitz constant of $x_t \mapsto d_\theta(\Pi_d(x_t))$. It was measured numerically by evaluating the Lipschitz constant for each of the $12$ constraints across $1024$ samples at every timestep, then taking the maximum value.}
    \label{fig:lipschitz}
\end{figure}

Figure~\ref{fig:lipschitz} shows how the Lipschitz constant $L_t$ evolves over diffusion time, measured by taking the maximum constant from all the constraints and sampling trajectories of \textsc{Data2D}. The figure supports the assumption that $x_t \mapsto d_\theta(\Pi_d(x_t))$ is $L_t$-Lipschitz and that $\alpha_t L_t < 1$ as a sensible one. Here $\alpha_t=1$.

\section{Detailed results} \label{app:detailed-results}

In this section we provide the detailed numerical results of the experiments from Section~\ref{sec:experiment} in the main text. Table~\ref{tab:ks_summary} extends Table~\ref{tab:ks-main-text} from the main text with the number of calls to the denoiser and includes other simpler constraints (Appendix~\ref{app:ks-additional}) in the statistics. Tables~\ref{tab:appa_2m_temperature}, \ref{tab:appa_10m_u_component_of_wind}, \ref{tab:appa_10m_v_component_of_wind}, \ref{tab:appa_mean_sea_level_pressure}, and~\ref{tab:appa_total_precipitation} give the average over time of the values in Figure~\ref{fig:appa}.

\input{tables/big_table}
\input{tables/appa_tables}

\FloatBarrier

\section{Additional Ablations} \label{app:abl}

\begin{figure}[!h]
    \centering
    \begin{subfigure}[t]{0.64\linewidth}
        \centering
        \includegraphics[width=\linewidth]{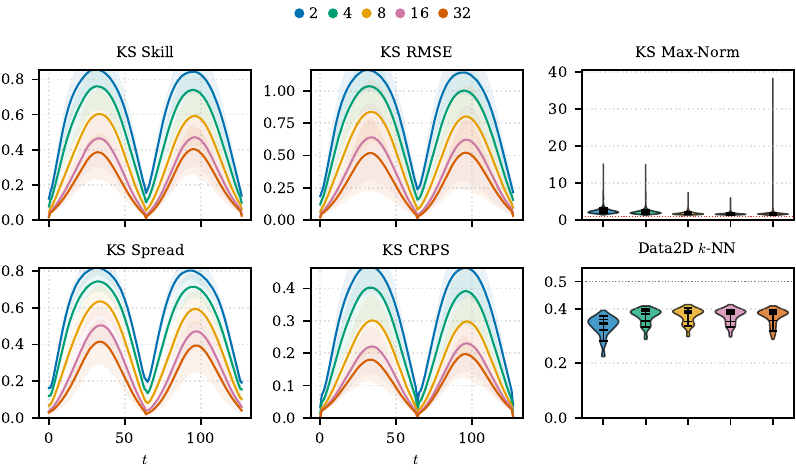}
        \caption{Metrics.}
        \label{fig:app-proj-metrics}
    \end{subfigure}\hfill
    \begin{subfigure}[t]{0.34\linewidth}
        \centering
        \includegraphics[width=\linewidth]{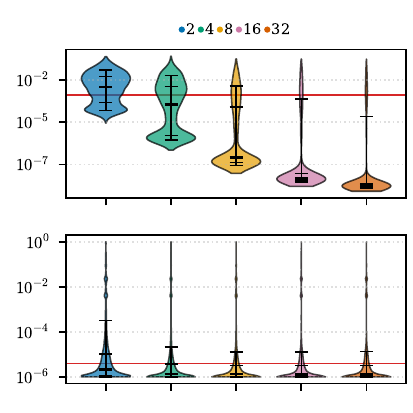}
        \caption{Constraint violation.}
        \label{fig:app-proj-viol}
    \end{subfigure}
    \caption{Projection-step ablation on \textsc{KS} and \textsc{Data2D}.  The
    labels $2,4,8,16,32$ denote the number of projection iterations used inside
    $\Pi_d$.  More projection iterations improve the local projection solve and
    reduce violation in both settings.  The gains are clearest on \textsc{KS};
    on \textsc{Data2D}, the $k$-NN cross-edge rate improves quickly and then
    starts to saturate, consistent with the lower-dimensional projection problem
    reaching a good local minimum after relatively few iterations.}
    \label{fig:app-proj}
\end{figure}

\begin{figure}[!h]
    \centering
    \begin{subfigure}[t]{0.64\linewidth}
        \centering
        \includegraphics[width=\linewidth]{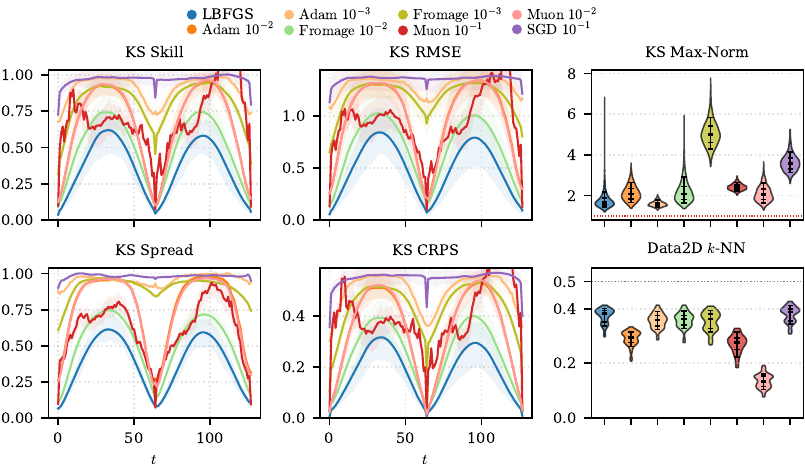}
        \caption{Metrics.}
        \label{fig:app-optimizer-metrics}
    \end{subfigure}\hfill
    \begin{subfigure}[t]{0.34\linewidth}
        \centering
        \includegraphics[width=\linewidth]{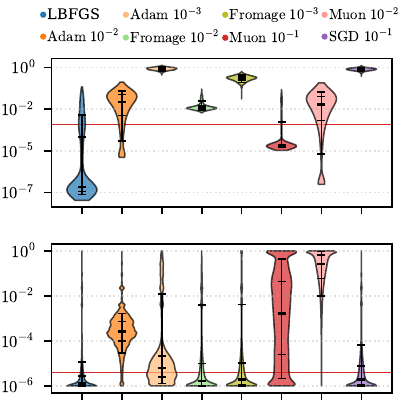}
        \caption{Constraint violation.}
        \label{fig:app-optimizer-viol}
    \end{subfigure}
    \caption{Optimizer ablation for the denoiser-space projection problem.  We
    compare L-BFGS, Adam, Fromage, Muon, and SGD with representative learning
    rates; divergent learning rates have been omitted.  L-BFGS is the
    strongest and most stable optimizer across metrics and violation, while the
    first-order methods are more sensitive to step size and generally give worse
    constrained distributions or less regular \textsc{KS} trajectories.}
    \label{fig:app-optimizer}
\end{figure}

\begin{figure}
    \centering
    \begin{subfigure}[t]{0.64\linewidth}
        \centering
        \includegraphics[width=\linewidth]{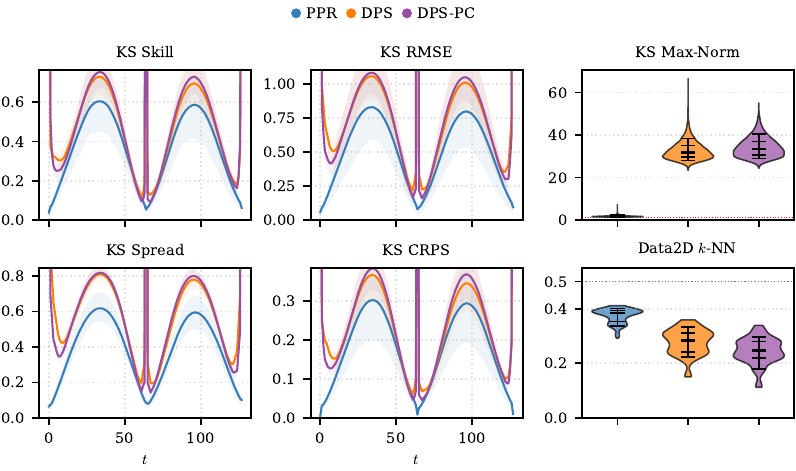}
        \caption{Metrics.}
        \label{fig:app-dps-pc-metrics}
    \end{subfigure}\hfill
    \begin{subfigure}[t]{0.34\linewidth}
        \centering
        \includegraphics[width=\linewidth]{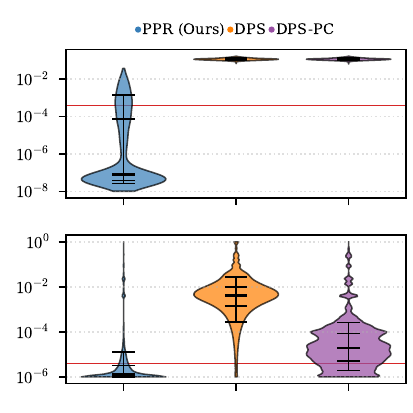}
        \caption{Constraint violation.}
        \label{fig:app-dps-pc-viol}
    \end{subfigure}
    \caption{DPS-PC ablation.  DPS-PC adds DPS-style correction steps in a
    predictor-corrector scheme.  Although this can improve some
    feasibility-related behavior relative to DPS, it does not close the gap to
    PPR.  The \textsc{KS} panels show that DPS-PC does not recover the accuracy
    and regularity obtained by projecting through the denoiser and renoising,
    and PPR remains the most reliable method for low-violation samples that
    preserve the data distribution.}
    \label{fig:app-dps-pc}
\end{figure}

This appendix reports three additional experiments that complement the
main-text ablation and baselines comparison: the number of projection steps
$\Pi_d(x_t) \equiv \textsc{PROJECT} = \arg\min_x c(d_\theta(x))$ (Figure~\ref{fig:app-proj} and Table~\ref{tab:ablation_proj}), the optimizer used for this inner
projection problem (Figure~\ref{fig:app-optimizer}), and a DPS predictor-corrector variant (DPS-PC) (Figure~\ref{fig:app-dps-pc}).

\input{tables/appendix_e_tables}

\FloatBarrier

\section{Additional \textsc{KS} Constraints} \label{app:ks-additional}

This section repeats the \textsc{KS} experiments on three additional
constraints: two easier observation constraints and a conservation law.  The main text uses the multi-time sine constraint
\[
    c(u_0)=\sum_i \|\sin(u_0^i)-\sin(u^i)\|^2,
    \qquad i\in\{0,64,127\},
\]
which constrains nonlinear observations at three system times.  Here we also
consider the single-time sine constraint
$c(u_0)=\|\sin(u_0^0)-\sin(u^0)\|^2$ and the linear observation
$c(u_0)=\|u_0^0-u^0\|^2$.  These alternatives are substantially less
challenging, especially the linear observation.
Consequently, the baselines can appear more competitive on the
metrics for these constraints. The relevant distinction is whether a method
also maintains feasibility and continuity; PPR remains the most reliable method
under that stricter criterion. The linear observation is forecasting from an initial condition, but it is not the main benchmark and we do not claim parity with specialized PDE forecasters.

Each figure below aggregates the two additional observation constraints, with the
single-time sine observation on the left and the linear observation on the
right.  Within each block, the figure reports constraint violation, Max-Norm,
spread, skill, RMSE, and CRPS.  In all metric panels, the x-axis is
the dynamical system time, not diffusion time.

\paragraph{Conservation of the spatial mean.} We also enforce a conservation law. The spatial mean $m^i = \frac{1}{L}\int_0^L u^i(x)\,dx$, with $L = 64$, is zero for our initial conditions and should remain zero (numerically $\sim 10^{-7}$) at all system times. We frame $m_0^i = 0$ for all $i$ as a constraint. For ease of comparison, we set the feasibility threshold to the largest violation observed among PPR samples. Therefore, even the worst PPR sample has a smaller violation than $(100-\text{feasibility})\%$ of the baseline samples. The unconstrained prior has $0\%$ feasibility (Table~\ref{tab:ks-mean}), showing that the pretrained model does not conserve the spatial mean on its own. PPR also achieves the lowest residual among all samplers, making it the closest to the fine-resolution GT and, under this numerical criterion, more feasible than the coarse GT the prior was trained on.

\paragraph{DiffusionPDE.} We run DiffusionPDE \citep{huang2024DiffusionPDE} with the same pretrained prior and select its two guidance weights per constraint with a grid search on feasibility at $1024$ diffusion steps. The row in Table~\ref{tab:ks-main-text} uses $4096$ steps and takes $91.5$\,s, comparable to PPR's $80.2$\,s, evaluated on the first $32$ test trajectories of the main constraint. At $1024$ steps ($25.8$\,s) on all $256$ test trajectories, feasibility is $9.1\%$ with RMSE $1.036$.

\input{tables/ks_mean_conservation}

\begin{figure}[t]
    \centering
    \includegraphics[width=\linewidth]{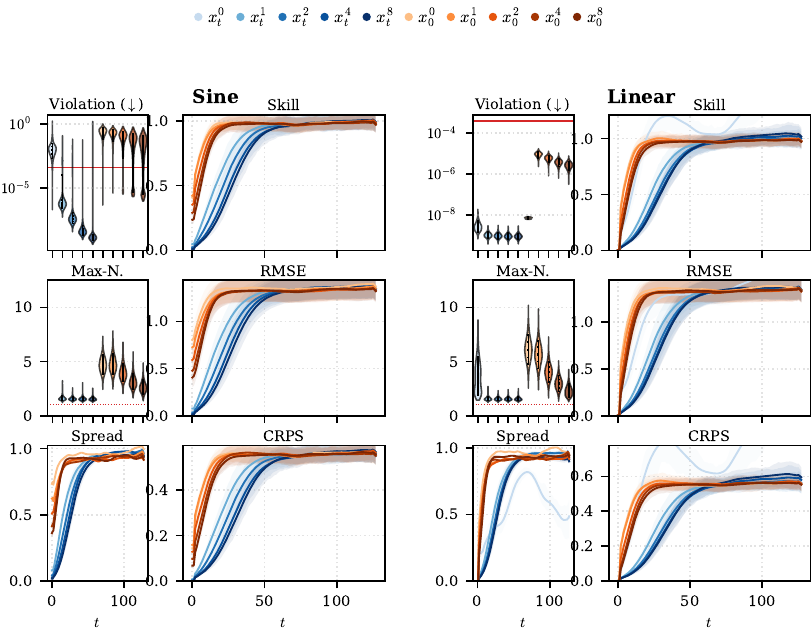}
    \caption{\textsc{KS} PPR ablation on additional constraints. Despite the easier
    single-time sine and linear observations, the same conclusion holds:
    projecting through the denoiser and renoising are necessary to preserve
    regular trajectories while reducing constraint violation. These easier
    constraints are less discriminating than the main-text multi-time sine
    constraint, but they still show the same failure mode for projections that
    do not preserve the data manifold.}
    \label{fig:app-ks-thm}
\end{figure}

\begin{figure}[t]
    \centering
    \includegraphics[width=\linewidth]{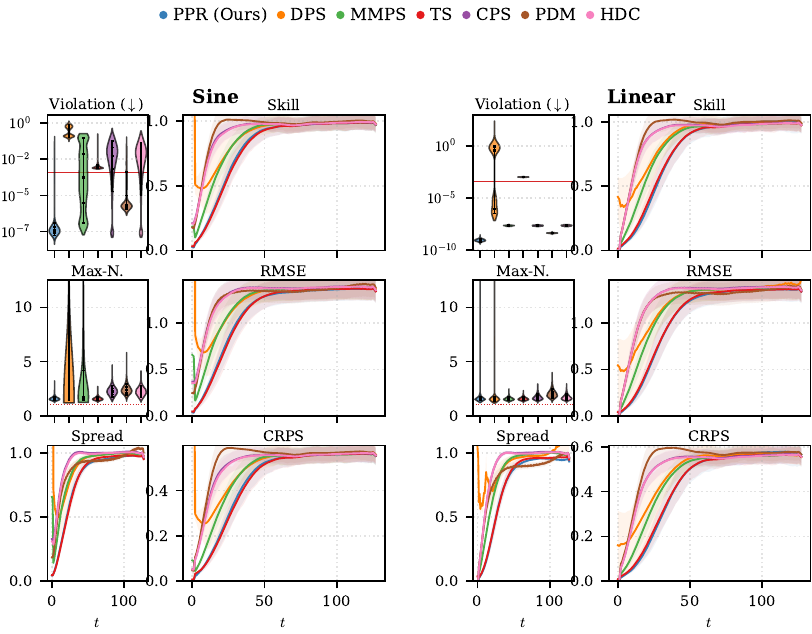}
    \caption{\textsc{KS} baseline comparison on additional constraints.  Because
    the single-time sine and linear observations are easier, several baselines
    become more competitive on the trajectory proxy metrics.  However, PPR
    remains the most consistent method when metrics are considered together with
    the constraint violation.}
    \label{fig:app-ks-eval}
\end{figure}

\begin{figure}[t]
    \centering
    \includegraphics[width=\linewidth]{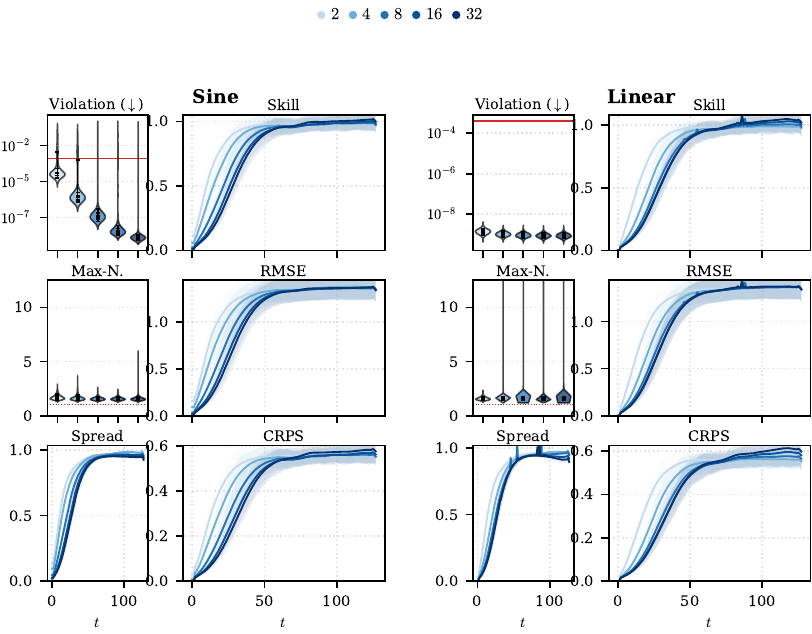}
    \caption{\textsc{KS} projection-step ablation on additional constraints.
    Increasing the number of projection iterations improves the local solution
    of $\Pi_d$ and generally lowers violation.}
    \label{fig:app-ks-proj}
\end{figure}

\begin{figure}[t]
    \centering
    \includegraphics[width=\linewidth]{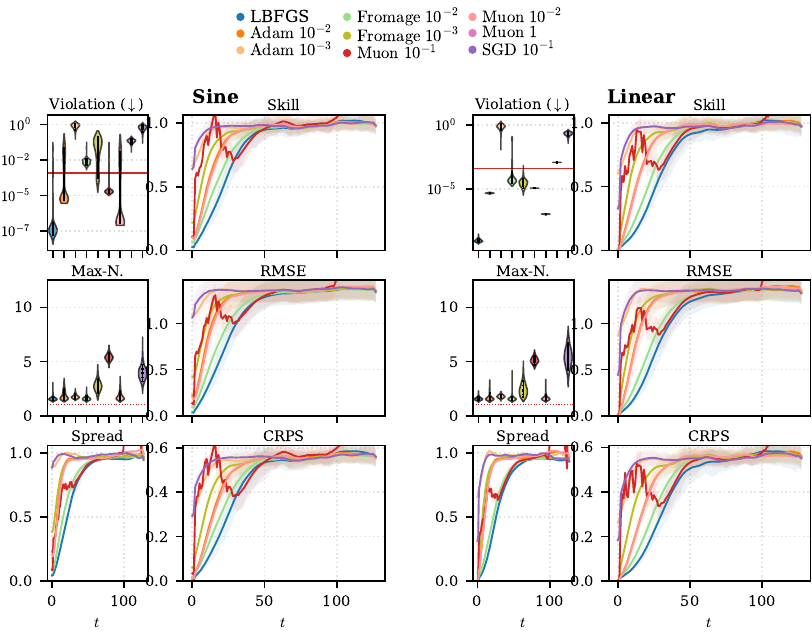}
    \caption{\textsc{KS} optimizer ablation on additional constraints. L-BFGS remains the most stable option
    across constraints, while first-order optimizers are more sensitive to their
    learning rates and less reliable in maintaining both feasibility and
    trajectory regularity.}
    \label{fig:app-ks-optimizer}
\end{figure}

\begin{figure}[t]
    \centering
    \includegraphics[width=\linewidth]{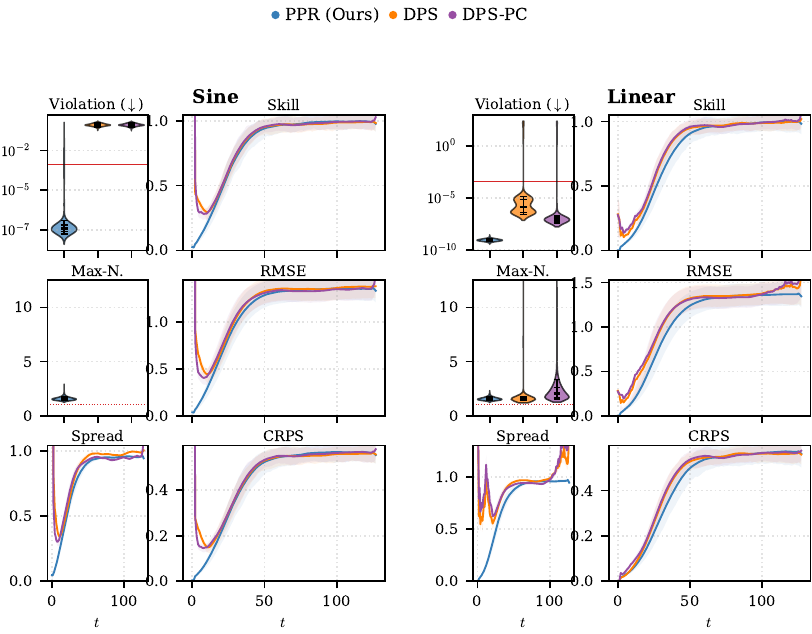}
    \caption{\textsc{KS} DPS-PC ablation on additional constraints.  DPS-PC
    improves over DPS in some easier cases, but this does not close the gap to
    PPR.}
    \label{fig:app-ks-dps-pc}
\end{figure}

\FloatBarrier

\FloatBarrier

\section{Example samples from all experiments} \label{app:samples}

We provide below some representative samples from each of the experiments. For \textsc{KS} and \textsc{Weather}, we simply picked a random ensemble member from each method to plot. Higher resolution versions are available in the supplementary material.

\begin{figure}
    \centering
    \includegraphics{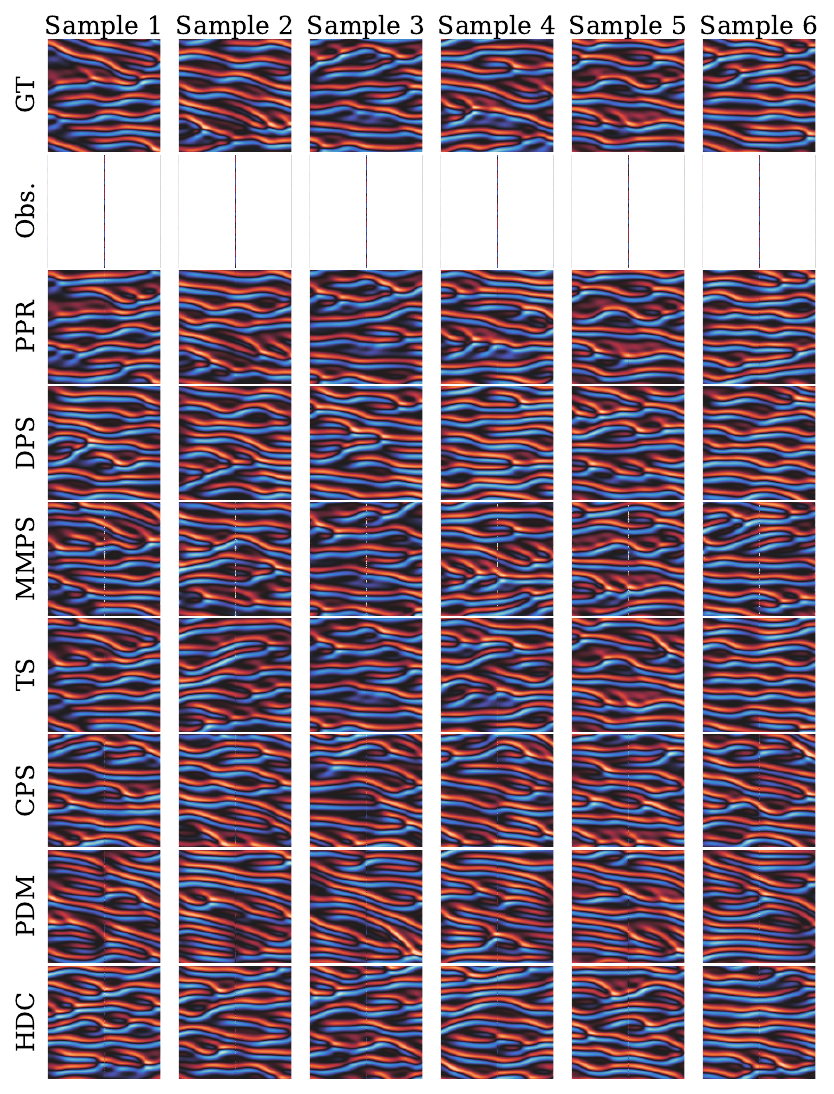}
    \caption{Samples from the sine observation at 3 different times, corresponding to the experiment in Section~\ref{sec:experiment}. For each sample from the ground truth set, we show one member of the ensemble produced by each method. The $y$-axis is the spatial domain, while the $x$-axis is the temporal domain.}
    \label{fig:app-ks-sin2}
\end{figure}

\begin{figure}
    \centering
    \includegraphics{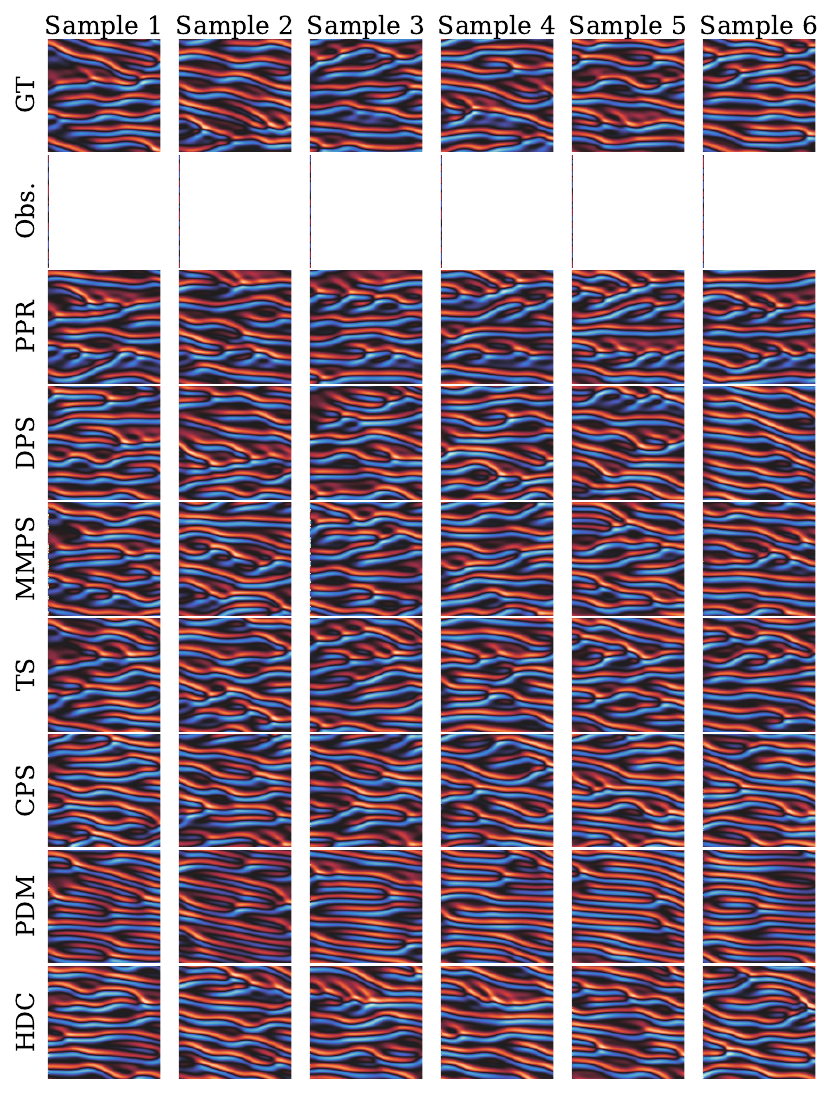}
    \caption{Samples from the sine observation at $t=0$. For each sample from the ground truth set, we show one member of the ensemble produced by each method. The $y$-axis is the spatial domain, while the $x$-axis is the temporal domain.}
    \label{fig:app-ks-sin}
\end{figure}

\begin{figure}
    \centering
    \includegraphics{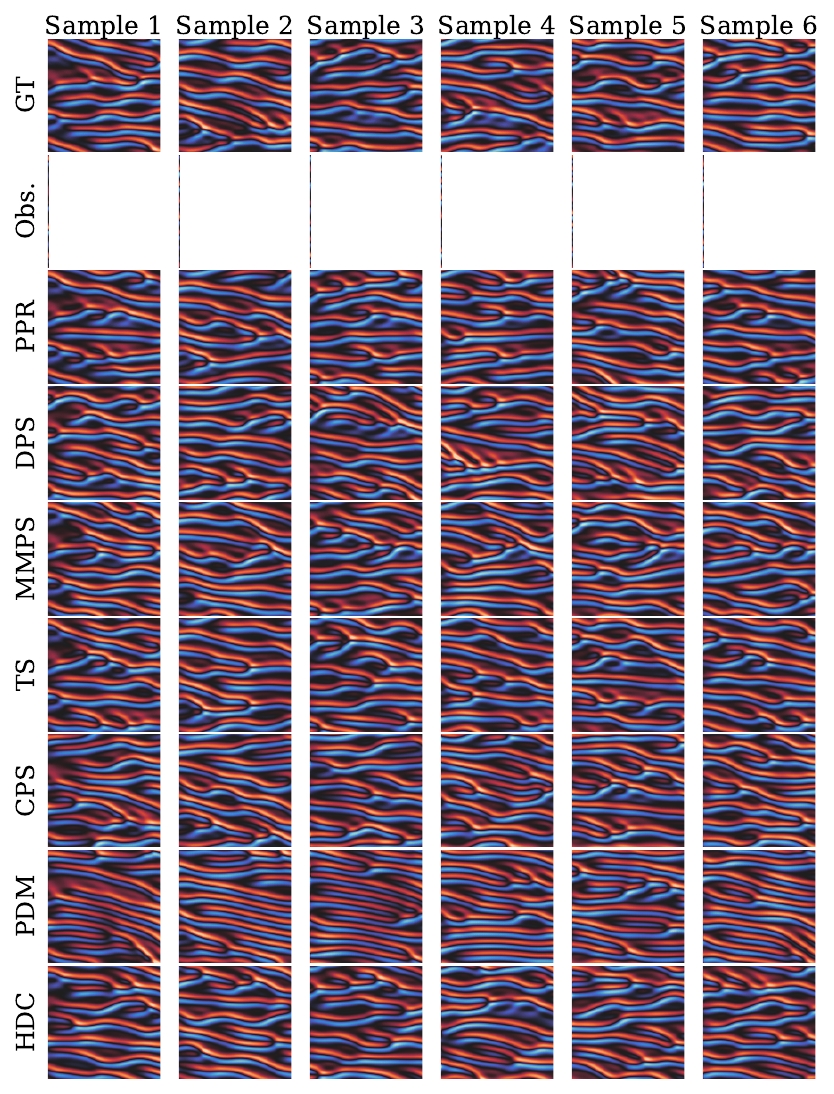}
    \caption{Samples from the linear observation at $t=0$. For each sample from the ground truth set, we show one member of the ensemble produced by each method. The $y$-axis is the spatial domain, while the $x$-axis is the temporal domain.}
    \label{fig:app-ks-linear}
\end{figure}

\FloatBarrier

\begin{figure}
    \centering
    \includegraphics{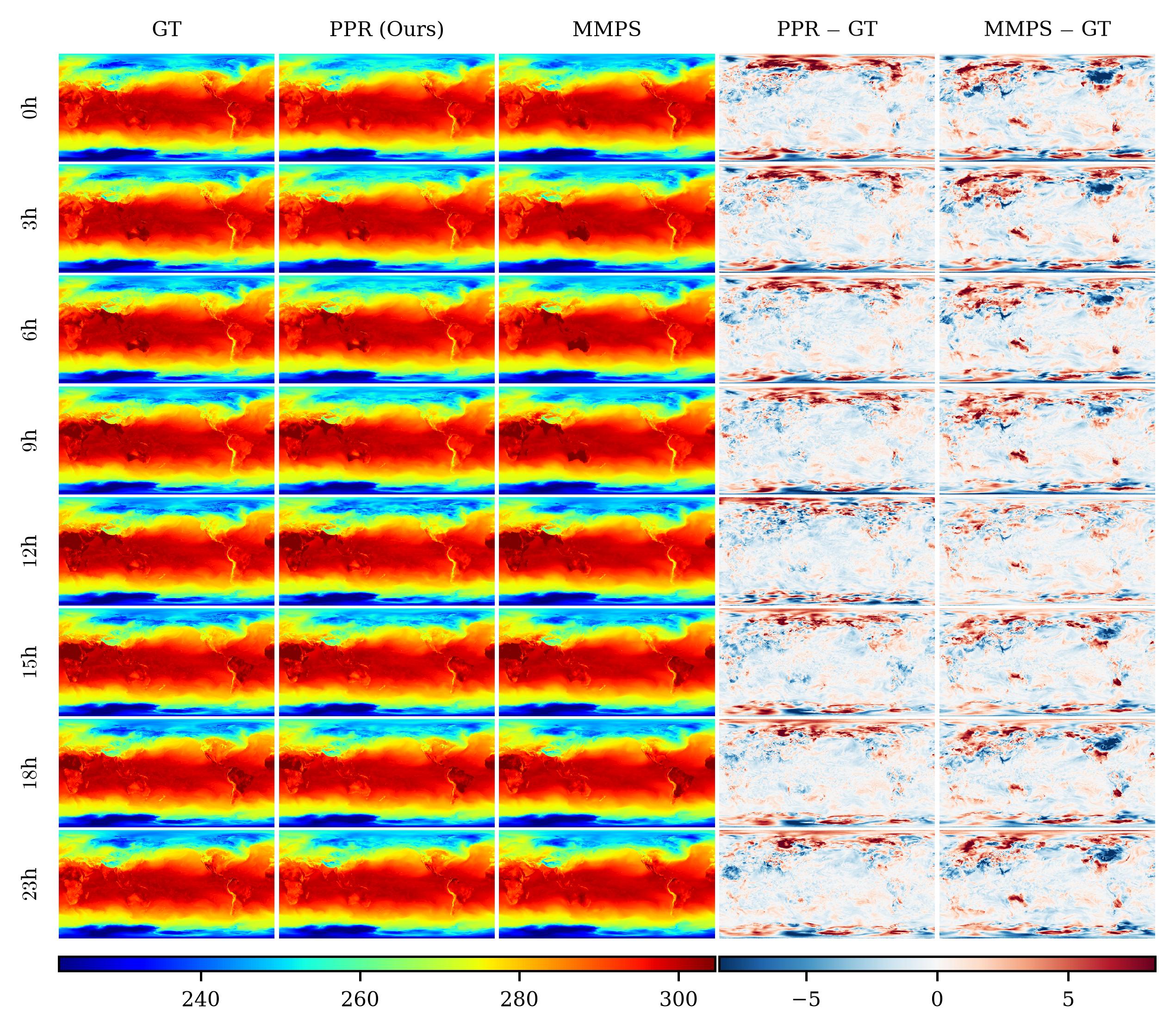}
    \caption{\textsc{Weather} sample showing the $2$m temperature (K) for the ground truth, PPR, and MMPS, and the errors of PPR and MMPS against the ground truth. The $y$-axis is the hour in $24$ h format.}
    \label{fig:app-2m-temp}
\end{figure}

\begin{figure}
    \centering
    \includegraphics{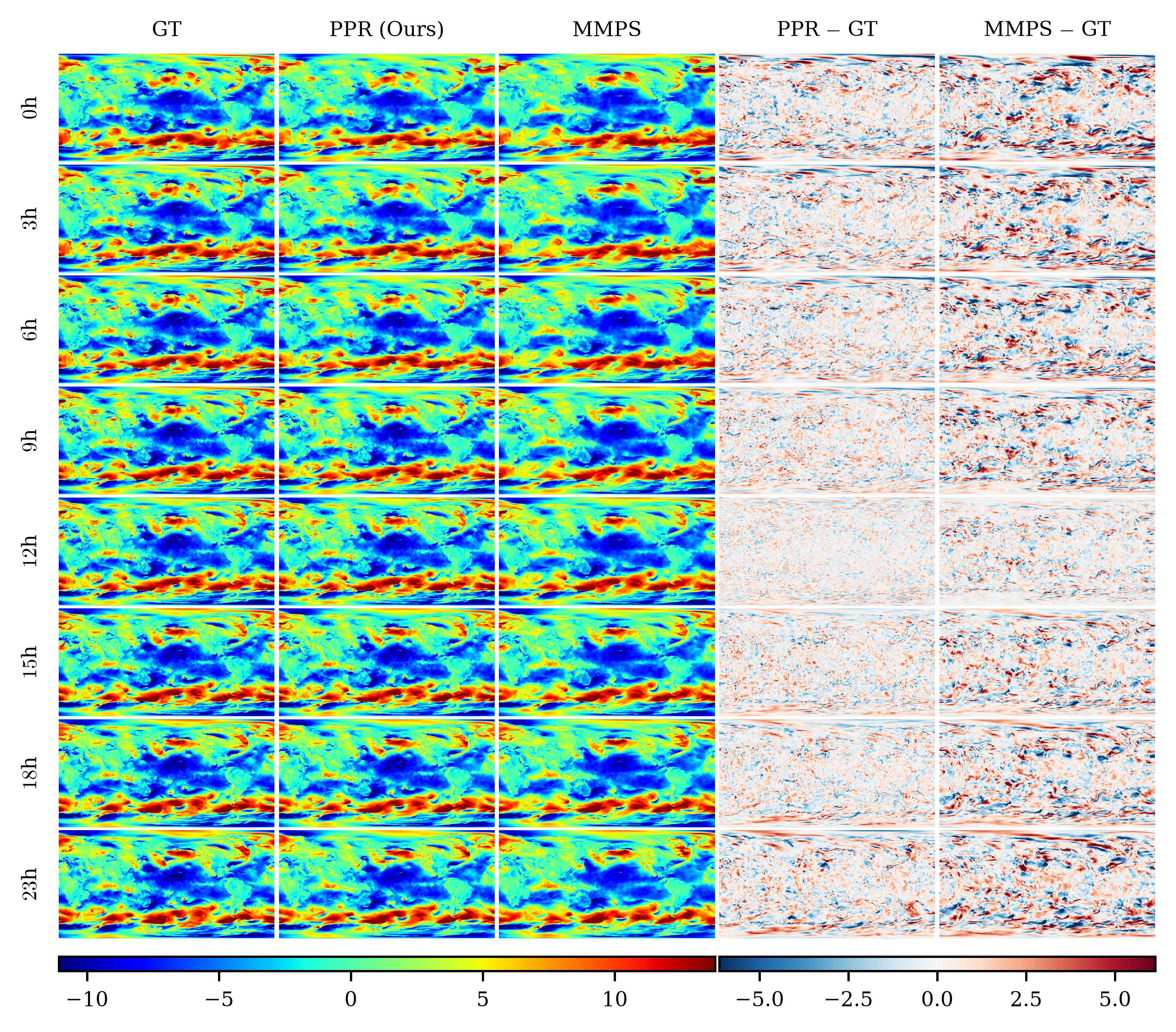}
    \caption{\textsc{Weather} sample showing the $10$m zonal wind component $u$ ($\frac{m}{s}$) for the ground truth, PPR, and MMPS, and the errors of PPR and MMPS against the ground truth. The $y$-axis is the hour in $24$ h format.}
    \label{fig:app-10m-u}
\end{figure}

\begin{figure}
    \centering
    \includegraphics{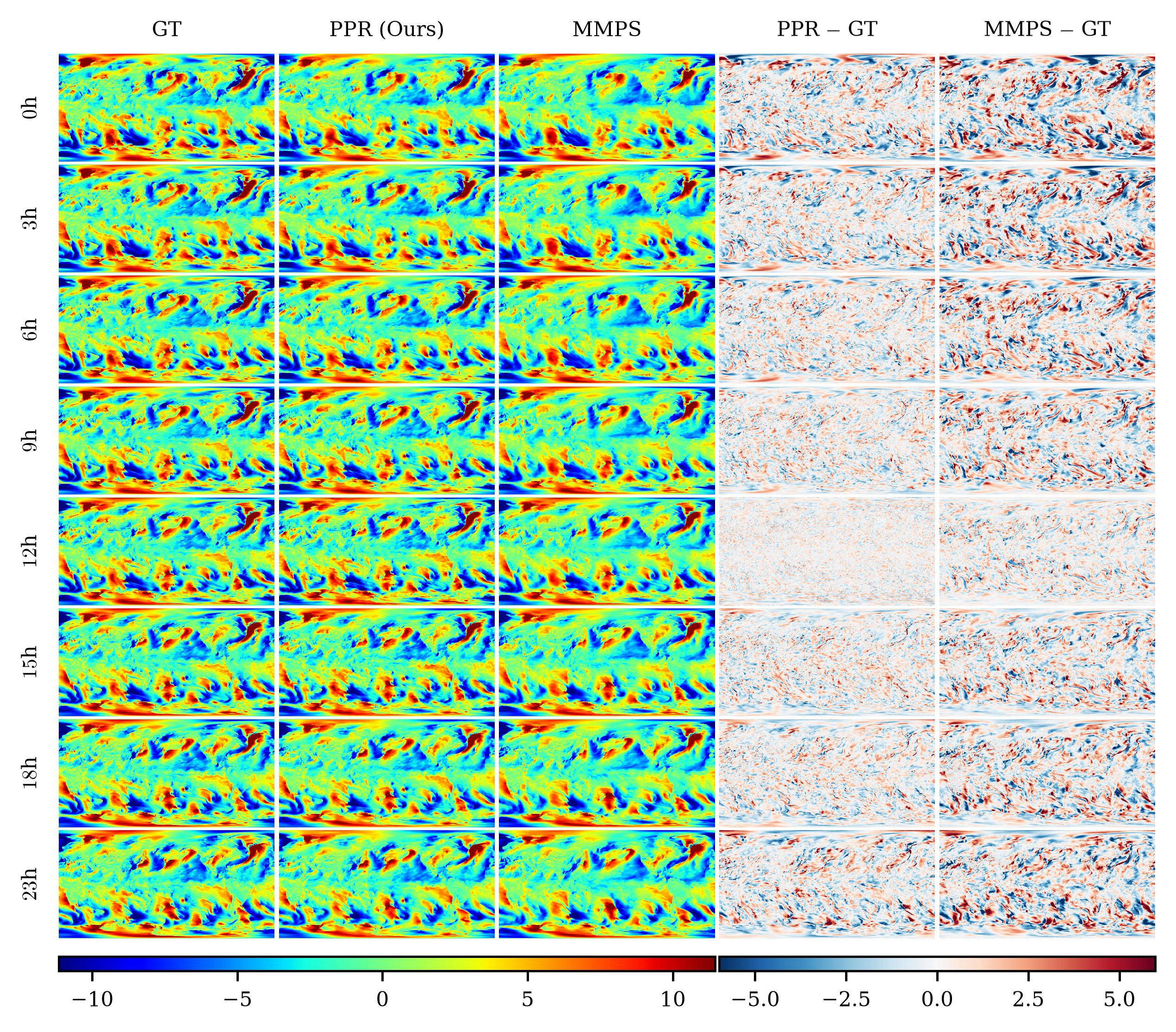}
    \caption{\textsc{Weather} sample showing the $10$m meridional wind component $v$ ($\frac{m}{s}$) for the ground truth, PPR, and MMPS, and the errors of PPR and MMPS against the ground truth. The $y$-axis is the hour in $24$ h format.}
    \label{fig:app-10m-v}
\end{figure}

\begin{figure}
    \centering
    \includegraphics{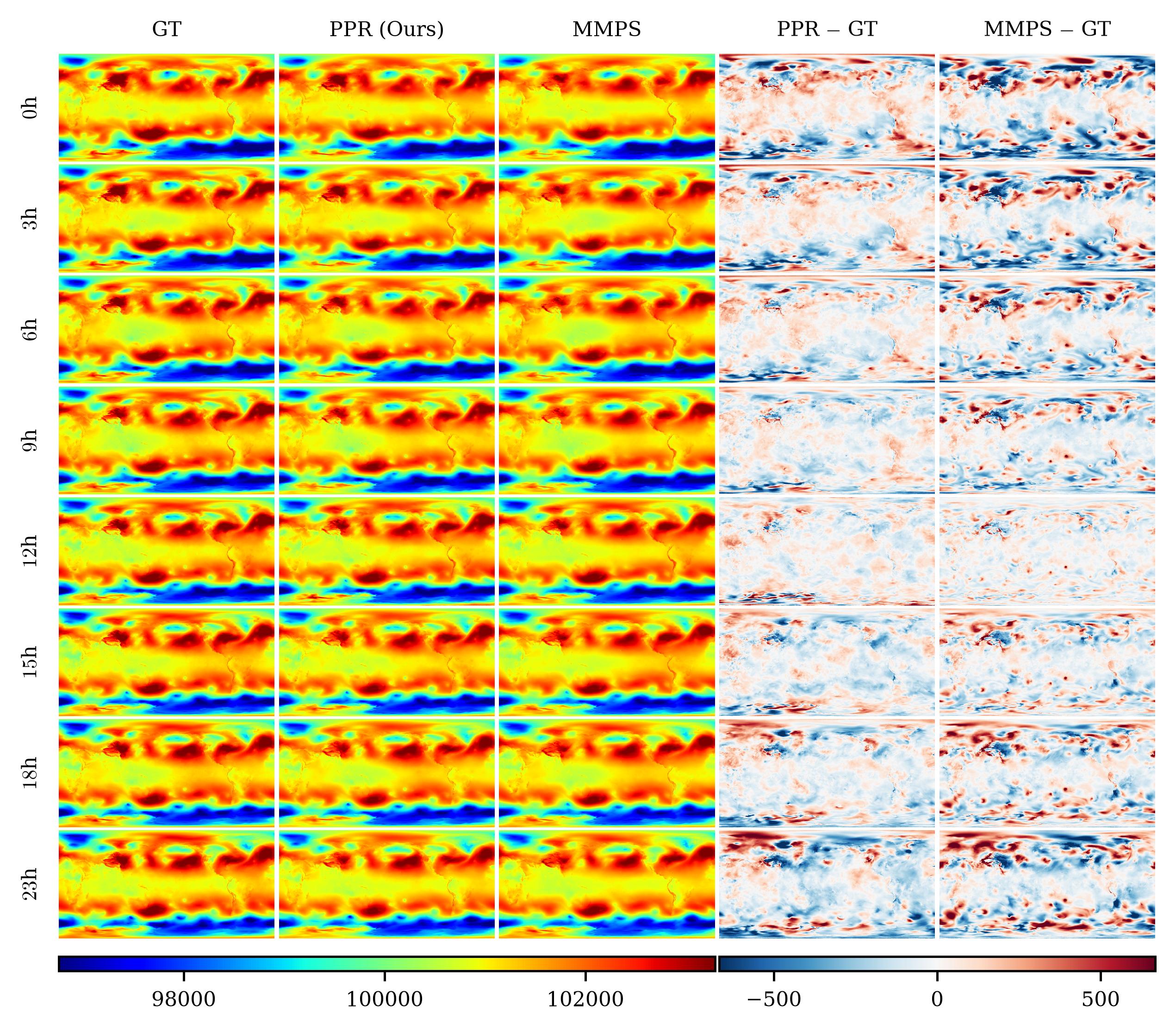}
    \caption{\textsc{Weather} sample showing the mean sea level pressure (Pa) for the ground truth, PPR, and MMPS, and the errors of PPR and MMPS against the ground truth. The $y$-axis is the hour in $24$ h format.}
    \label{fig:app-mslp}
\end{figure}

\begin{figure}
    \centering
    \includegraphics{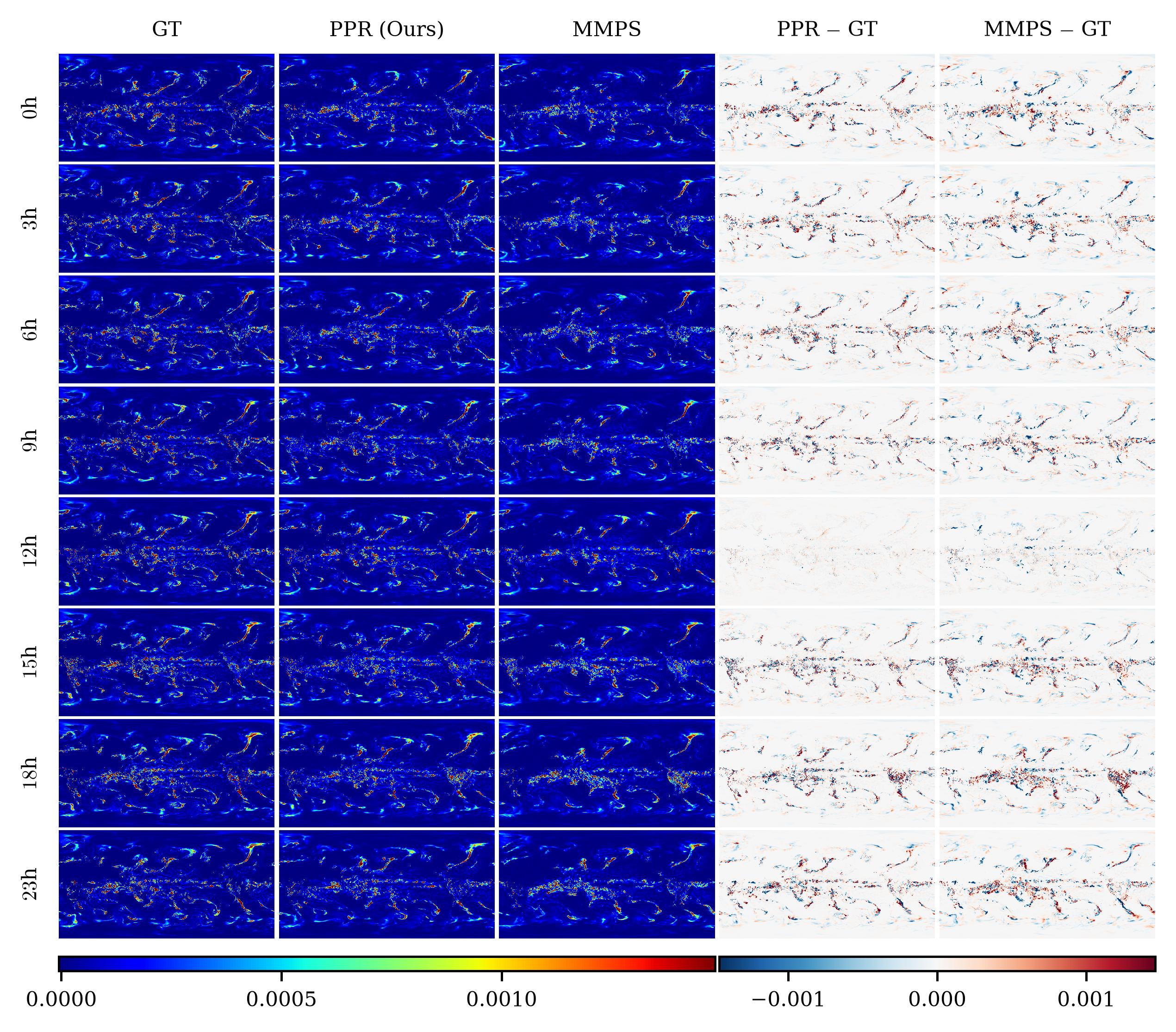}
    \caption{\textsc{Weather} sample showing the total precipitation (m) for the ground truth, PPR, and MMPS, and the errors of PPR and MMPS against the ground truth. The $y$-axis is the hour in $24$ h format.}
    \label{fig:app-tp}
\end{figure}

\end{document}

%% file: tables/arch_table.tex
\begin{table}[h] 
\centering
\caption{Architecture and noise-schedule parameters for KS and Data2D.}
\label{tab:architectures}
\begin{tabular}{lcc}
\toprule
 & \textbf{Data2D} & \textbf{KS} \\
\midrule
Architecture        & Mod MLP          & UNet \\
Depth / channels    & 3 layers, 256 units   & {[32, 64, 128, 256]} \\
Activation          & SiLU                  & SiLU \\
Time conditioning   & FiLM (per layer)      & FiLM (per layer) \\
Kernel size         & ---                   & $3 \times 3$ \\
Spatial padding     & ---                   & circular \\
\midrule
Noise schedule      & \multicolumn{2}{c}{log logit} \\
$\sigma_{\min}$     & $10^{-3}$             & $10^{-4}$ \\
$\sigma_{\max}$     & $10^{2}$              & $10^{4}$ \\
Spread $s$              & 2.0                   & 4.0 \\
\bottomrule
\end{tabular}
\end{table}

%% file: tables/big_table.tex
\begin{table}[t]
\centering
\caption{%
    KS results. T: diffusion steps. fwd is the number of network call. bwd is the number of gradient calls (adds one to fwd too). The feasibility shows the \% of samples under a threshold (0 is worse, 100 is the best). Max-Norm is better close to $1$. Note that the feasibility is averaged over different constraints. For the most challenging constraints, the feasibility of PPR is over $84\%$ compared to $8\%$ for the next best method (main text).
}
\label{tab:ks_summary}
\tiny
\setlength{\tabcolsep}{4pt}
\begin{tabular}{l c r r r r r r r r r}
\toprule
Method & $T$ & fwd & bwd & Time (s) & Feas.\,$\uparrow$ & Skill\,$\downarrow$ & RMSE\,$\downarrow$ & CRPS\,$\downarrow$ & Spread\,$\downarrow$ & MaxNorm \\
\midrule
    PPR (Ours)           & 64 &         1216 &         1024 & 84.8$\,\pm\,$8.1   & \textbf{86.9\%}      & \textbf{0.648$\,\pm\,$0.212} & \textbf{0.891$\,\pm\,$0.285} & \textbf{0.351$\,\pm\,$0.130} & \textbf{0.644$\,\pm\,$0.189} & \textbf{1.65$\,\pm\,$0.17} \\
    DPS                  & 1024 &         2048 &         1024 & 25.6$\,\pm\,$0.3   & 11.0\%               & 0.817$\,\pm\,$0.153 & 1.240$\,\pm\,$0.165 & 0.450$\,\pm\,$0.099 & 0.956$\,\pm\,$0.129 & 16.64$\,\pm\,$22.82 \\
    MMPS                 & 1024 &         5120 &        20480 & 306$\,\pm\,$3      & 42.1\%               & 0.735$\,\pm\,$0.161 & 1.069$\,\pm\,$0.179 & 0.399$\,\pm\,$0.101 & 0.801$\,\pm\,$0.101 & 9.08$\,\pm\,$10.30 \\
    Trust                & 192 &        24960 &        12288 & 405$\,\pm\,$11     & 0.0\%                & 0.667$\,\pm\,$0.181 & 0.929$\,\pm\,$0.233 & 0.358$\,\pm\,$0.114 & 0.682$\,\pm\,$0.144 & 1.71$\,\pm\,$0.25 \\
    CPS                  & 1024 &         1024 &            0 & 52.6$\,\pm\,$3.1   & 36.0\%               & 0.873$\,\pm\,$0.062 & 1.242$\,\pm\,$0.055 & 0.483$\,\pm\,$0.042 & 0.911$\,\pm\,$0.022 & 2.82$\,\pm\,$1.22 \\
    PDM                  & 384 &        24576 &            0 & 296$\,\pm\,$3      & 56.9\%               & 0.897$\,\pm\,$0.064 & 1.211$\,\pm\,$0.066 & 0.508$\,\pm\,$0.043 & 0.837$\,\pm\,$0.045 & 2.97$\,\pm\,$1.10 \\
    HDC                  & 1024 &         1024 &            0 & 49.5$\,\pm\,$0.9   & 36.3\%               & 0.865$\,\pm\,$0.068 & 1.229$\,\pm\,$0.063 & 0.478$\,\pm\,$0.046 & 0.902$\,\pm\,$0.027 & 2.76$\,\pm\,$1.16 \\
\bottomrule
\end{tabular}
\end{table}

%% file: tables/appa_tables.tex
\begin{table}[t]
\centering
\caption{Appa results for 2m temperature [K]. Time is the wallclock to produce 32 samples sequentially. All values are mean\,$\pm$\,std across the 24 timesteps.}
\label{tab:appa_2m_temperature}
\setlength{\tabcolsep}{4pt}
\begin{tabular}{l c r r r r}
\toprule
Method & Time & Skill\,$\downarrow$ & CRPS\,$\downarrow$ & Spread\,$\downarrow$ & Spread/Skill \\
\midrule
    PPR (Ours) & 2h 01m & \textbf{1.399$\,\pm\,$0.080} & \textbf{0.693$\,\pm\,$0.045} & \textbf{0.807$\,\pm\,$0.069} & 0.589$\,\pm\,$0.072 \\
    MMPS & 2h 05m & 1.536$\,\pm\,$0.134 & 0.708$\,\pm\,$0.055 & 1.129$\,\pm\,$0.102 & \textbf{0.746$\,\pm\,$0.009} \\
    TS & 1h 53m & 3.525$\,\pm\,$0.295 & 1.564$\,\pm\,$0.105 & 2.238$\,\pm\,$0.109 & 0.648$\,\pm\,$0.041 \\
    CPS & 1h 26m & 3.786$\,\pm\,$0.104 & 1.675$\,\pm\,$0.040 & 2.274$\,\pm\,$0.077 & 0.610$\,\pm\,$0.017 \\
    HDC & 1h 26m & 2.996$\,\pm\,$0.053 & 1.385$\,\pm\,$0.024 & 1.562$\,\pm\,$0.019 & 0.530$\,\pm\,$0.012 \\
    DPS & 2h 15m & 4.918$\,\pm\,$0.542 & 2.165$\,\pm\,$0.196 & 2.892$\,\pm\,$0.126 & 0.605$\,\pm\,$0.072 \\
\bottomrule
\end{tabular}
\end{table}

\begin{table}[t]
\centering
\caption{Appa results for 10m U-wind component [m/s]. Time is the wallclock to produce 32 samples sequentially. All values are mean\,$\pm$\,std across the 24 timesteps.}
\label{tab:appa_10m_u_component_of_wind}
\setlength{\tabcolsep}{4pt}
\begin{tabular}{l c r r r r}
\toprule
Method & Time & Skill\,$\downarrow$ & CRPS\,$\downarrow$ & Spread\,$\downarrow$ & Spread/Skill \\
\midrule
    PPR (Ours) & 2h 01m & \textbf{1.104$\,\pm\,$0.181} & \textbf{0.599$\,\pm\,$0.095} & \textbf{0.755$\,\pm\,$0.053} & 0.715$\,\pm\,$0.136 \\
    MMPS & 2h 05m & 1.581$\,\pm\,$0.236 & 0.796$\,\pm\,$0.108 & 1.293$\,\pm\,$0.203 & 0.829$\,\pm\,$0.021 \\
    TS & 1h 53m & 3.718$\,\pm\,$0.374 & 1.923$\,\pm\,$0.194 & 2.858$\,\pm\,$0.222 & 0.784$\,\pm\,$0.044 \\
    CPS & 1h 26m & 4.122$\,\pm\,$0.067 & 2.129$\,\pm\,$0.041 & 3.002$\,\pm\,$0.108 & 0.740$\,\pm\,$0.026 \\
    HDC & 1h 26m & 4.075$\,\pm\,$0.065 & 2.117$\,\pm\,$0.039 & 2.518$\,\pm\,$0.033 & 0.628$\,\pm\,$0.011 \\
    DPS & 2h 15m & 3.923$\,\pm\,$0.461 & 2.072$\,\pm\,$0.244 & 3.537$\,\pm\,$0.281 & \textbf{0.924$\,\pm\,$0.077} \\
\bottomrule
\end{tabular}
\end{table}

\begin{table}[t]
\centering
\caption{Appa results for 10m V-wind component [m/s]. Time is the wallclock to produce 32 samples sequentially. All values are mean\,$\pm$\,std across the 24 timesteps.}
\label{tab:appa_10m_v_component_of_wind}
\setlength{\tabcolsep}{4pt}
\begin{tabular}{l c r r r r}
\toprule
Method & Time & Skill\,$\downarrow$ & CRPS\,$\downarrow$ & Spread\,$\downarrow$ & Spread/Skill \\
\midrule
    PPR (Ours) & 2h 01m & \textbf{1.098$\,\pm\,$0.224} & \textbf{0.594$\,\pm\,$0.117} & \textbf{0.770$\,\pm\,$0.061} & 0.741$\,\pm\,$0.160 \\
    MMPS & 2h 05m & 1.599$\,\pm\,$0.250 & 0.815$\,\pm\,$0.122 & 1.297$\,\pm\,$0.219 & 0.822$\,\pm\,$0.021 \\
    TS & 1h 53m & 3.722$\,\pm\,$0.391 & 1.907$\,\pm\,$0.196 & 2.910$\,\pm\,$0.230 & 0.798$\,\pm\,$0.046 \\
    CPS & 1h 26m & 4.137$\,\pm\,$0.058 & 2.114$\,\pm\,$0.033 & 3.042$\,\pm\,$0.110 & 0.747$\,\pm\,$0.025 \\
    HDC & 1h 26m & 4.140$\,\pm\,$0.066 & 2.159$\,\pm\,$0.045 & 2.389$\,\pm\,$0.055 & 0.586$\,\pm\,$0.015 \\
    DPS & 2h 15m & 3.854$\,\pm\,$0.460 & 2.006$\,\pm\,$0.233 & 3.570$\,\pm\,$0.286 & \textbf{0.949$\,\pm\,$0.079} \\
\bottomrule
\end{tabular}
\end{table}

\begin{table}[t]
\centering
\caption{Appa results for Mean sea level pressure [Pa]. Time is the wallclock to produce 32 samples sequentially. All values are mean\,$\pm$\,std across the 24 timesteps.}
\label{tab:appa_mean_sea_level_pressure}
\setlength{\tabcolsep}{4pt}
\begin{tabular}{l c r r r r}
\toprule
Method & Time & Skill\,$\downarrow$ & CRPS\,$\downarrow$ & Spread\,$\downarrow$ & Spread/Skill \\
\midrule
    PPR (Ours) & 2h 01m & \textbf{111.186$\,\pm\,$23.199} & \textbf{59.473$\,\pm\,$13.391} & \textbf{68.678$\,\pm\,$5.978} & 0.653$\,\pm\,$0.141 \\
    MMPS & 2h 05m & 141.364$\,\pm\,$32.115 & 69.818$\,\pm\,$15.414 & 117.677$\,\pm\,$27.700 & \textbf{0.843$\,\pm\,$0.024} \\
    TS & 1h 53m & 636.298$\,\pm\,$92.946 & 337.614$\,\pm\,$47.335 & 281.827$\,\pm\,$36.586 & 0.452$\,\pm\,$0.020 \\
    CPS & 1h 26m & 756.719$\,\pm\,$17.679 & 393.572$\,\pm\,$10.043 & 325.945$\,\pm\,$8.827 & 0.437$\,\pm\,$0.006 \\
    HDC & 1h 26m & 727.634$\,\pm\,$16.358 & 361.569$\,\pm\,$9.251 & 338.532$\,\pm\,$7.344 & 0.472$\,\pm\,$0.003 \\
    DPS & 2h 15m & 667.376$\,\pm\,$108.510 & 361.394$\,\pm\,$57.688 & 306.990$\,\pm\,$38.747 & 0.474$\,\pm\,$0.048 \\
\bottomrule
\end{tabular}
\end{table}

\begin{table}[t]
\centering
\caption{Appa results for Total precipitation [mm]. Time is the wallclock to produce 32 samples sequentially. All values are mean\,$\pm$\,std across the 24 timesteps.}
\label{tab:appa_total_precipitation}
\setlength{\tabcolsep}{4pt}
\begin{tabular}{l c r r r r}
\toprule
Method & Time & Skill\,$\downarrow$ & CRPS\,$\downarrow$ & Spread\,$\downarrow$ & Spread/Skill \\
\midrule
    PPR (Ours) & 2h 01m & \textbf{0.345$\,\pm\,$0.084} & \textbf{0.068$\,\pm\,$0.014} & \textbf{0.220$\,\pm\,$0.042} & 0.666$\,\pm\,$0.086 \\
    MMPS & 2h 05m & 0.371$\,\pm\,$0.066 & 0.075$\,\pm\,$0.011 & 0.269$\,\pm\,$0.047 & 0.737$\,\pm\,$0.029 \\
    TS & 1h 53m & 0.464$\,\pm\,$0.018 & 0.101$\,\pm\,$0.004 & 0.495$\,\pm\,$0.092 & \textbf{1.079$\,\pm\,$0.190} \\
    CPS & 1h 26m & 0.470$\,\pm\,$0.012 & 0.102$\,\pm\,$0.003 & 0.532$\,\pm\,$0.047 & 1.149$\,\pm\,$0.104 \\
    HDC & 1h 26m & 0.462$\,\pm\,$0.012 & 0.103$\,\pm\,$0.003 & 0.251$\,\pm\,$0.017 & 0.552$\,\pm\,$0.031 \\
    DPS & 2h 15m & 0.476$\,\pm\,$0.018 & 0.104$\,\pm\,$0.004 & 0.708$\,\pm\,$0.150 & 1.503$\,\pm\,$0.299 \\
\bottomrule
\end{tabular}
\end{table}

%% file: tables/appendix_e_tables.tex
\begin{table}[!h]
\centering
\small
\caption{Effect of the number of projection steps on PPR performance. Columns as in the main text.}
\label{tab:ablation_proj}
\resizebox{\linewidth}{!}{%
\begin{tabular}{c | c c c | c c c c c c}
\toprule
\multicolumn{1}{c|}{} & \multicolumn{3}{c|}{Data2D} & \multicolumn{6}{c}{KS} \\
\cmidrule(lr){2-4} \cmidrule(lr){5-10}
Projection steps & Feas.\,(\%) $\uparrow$ & KNN & Time\,(s) $\downarrow$ & Feas.\,(\%) $\uparrow$ & RMSE $\downarrow$ & CRPS $\downarrow$ & Skill $\downarrow$ & Max-Norm & Time\,(s) $\downarrow$ \\
\midrule
    2 & $62.0$ & $0.336 \pm 0.036$ & $9.4 \pm 0.4$ & $33.3$ & $1.140 \pm 0.171$ & $0.445 \pm 0.093$ & $0.825 \pm 0.133$ & $1.91 \pm 0.44$ & $39.1 \pm 2.1$ \\
    4 & $76.0$ & $0.371 \pm 0.027$ & $10.4 \pm 0.5$ & $73.2$ & $1.069 \pm 0.218$ & $0.403 \pm 0.111$ & $0.770 \pm 0.167$ & $1.79 \pm 0.31$ & $53.0 \pm 2.2$ \\
    8 & {\boldmath$78.5$} & {\boldmath$0.374 \pm 0.027$} & $14.1 \pm 0.8$ & $87.2$ & $1.012 \pm 0.605$ & $0.361 \pm 0.134$ & $0.711 \pm 0.229$ & $1.65 \pm 0.29$ & $83.7 \pm 6.6$ \\
    16 & $78.2$ & $0.373 \pm 0.027$ & $21.3 \pm 1.2$ & $94.0$ & $0.925 \pm 0.340$ & $0.337 \pm 0.150$ & $0.666 \pm 0.256$ & {\boldmath$1.60 \pm 0.21$} & $150 \pm 25$ \\
    32 & $78.0$ & $0.370 \pm 0.029$ & $35.3 \pm 2.0$ & {\boldmath$95.9$} & {\boldmath$0.904 \pm 0.440$} & {\boldmath$0.330 \pm 0.162$} & {\boldmath$0.652 \pm 0.285$} & $1.61 \pm 0.41$ & $305 \pm 97$ \\
\bottomrule
\end{tabular}%
}
\end{table}

%% file: tables/ks_mean_conservation.tex
\begin{table}[t]
\centering
\caption{Conservation of the spatial mean on \textsc{KS}. The feasibility threshold was set at $\max_i |m^i| < 4 \times 10^{-5}$. GT (fine) and GT (coarse) are the test trajectories at resolutions $512$ and $128$.}
\label{tab:ks-mean}

\begin{tabular}{l | c c c}
\toprule
Method & Time\,(s) $\downarrow$ & $\max_i |m^i|$ $\downarrow$ & Feas.\,(\%) $\uparrow$ \\
\midrule
    GT (fine) & --- & $(2.67 \pm 1.40) \times 10^{-7}$ & $100.0$ \\
    GT (coarse) & --- & $(2.14 \pm 0.45) \times 10^{-4}$ & $0.4$ \\
    Prior & --- & $(8.96 \pm 3.81) \times 10^{-2}$ & $0.0$ \\
\midrule
    PPR (Ours) & $127 \pm 2$ & $(2.99 \pm 0.59) \times 10^{-5}$ & {\boldmath$100.0$} \\
    DPS & $25.6 \pm 0.1$ & $(3.61 \pm 11.80) \times 10^{-2}$ & $0.0$ \\
    MMPS & $305 \pm 0$ & $(3.85 \pm 0.53) \times 10^{-5}$ & $68.8$ \\
    TS & $408 \pm 0$ & $(4.02 \pm 0.43) \times 10^{-5}$ & $43.8$ \\
    CPS & $79.1 \pm 0.5$ & $(3.68 \pm 0.39) \times 10^{-5}$ & $75.0$ \\
    PDM & $295 \pm 0$ & $(1.22 \pm 0.28) \times 10^{-3}$ & $0.0$ \\
    HDC & $113 \pm 2$ & $(3.78 \pm 0.41) \times 10^{-5}$ & $68.8$ \\
    DiffusionPDE & $27.0 \pm 2.3$ & $(5.09 \pm 2.17) \times 10^{-3}$ & $0.0$ \\
\bottomrule
\end{tabular}
\end{table}